\documentclass[journal,compsoc,dvipsnames,9pt]{IEEEtran}
\usepackage{amsmath,amsfonts}
\usepackage{algorithm}
\usepackage{algpseudocode}
\usepackage{enumitem}
\usepackage{array}
\usepackage{threeparttable}
\usepackage[caption=false,font=normalsize,labelfont=sf]{subfig}
\usepackage[labelsep=period]{caption} 
\usepackage{textcomp}
\usepackage{stfloats}
\usepackage{tikz}
\usepackage{orcidlink}
\usepackage{url}
\usepackage{pifont}
\usepackage{verbatim}
\usepackage{array} 
\usepackage{bm} 
\usepackage{graphicx}
\usepackage{cite}
\usepackage{booktabs} 
\usepackage{adjustbox} 
\usepackage{multirow}
\usepackage[table]{xcolor}
\usepackage{doi}
\begin{document}
\algblock{ParFor}{EndParFor}
\algnewcommand\algorithmicparfor{\textbf{parfor}}
\algnewcommand\algorithmicpardo{\textbf{do}}
\algnewcommand\algorithmicendparfor{\textbf{end\ parfor}}
\algrenewtext{ParFor}[1]{\algorithmicparfor\ #1\ \algorithmicpardo}
\algrenewtext{EndParFor}{\algorithmicendparfor}

\bstctlcite{IEEEexample:BSTcontrol}
    \title{Unsupervised Active Learning via Natural Feature Progressive Framework}
  \author{~Yuxi~Liu\orcidlink{0009-0007-3083-0426},~\IEEEmembership{{student Member,~IEEE,
        ~Catherine~Lalman,
       and~Yimin~Yang\orcidlink{0000-0002-1131-2056},~\IEEEmembership{Senior Member,~IEEE}}}
    \thanks{This work was supported in part by the Natural Sciences and Engineering Research Council of Canada (NSERC) Discovery Grant Program under Grant RGPIN-2025-06527 and in part by the Western Research Award.}
    \thanks{Y. Liu is with the Department of Electrical and Computer Engineering, Western University, London, Canada.}  
    \thanks{C. Lalman is with the Department of Pathology and Genomic Medicine, Thomas Jefferson University, Philadelphia, PA, USA.}   
    \thanks{Y. Yang is with the Department of Electrical and Computer Engineering, Western University, London, Canada, and also with the Vector Institute, Toronto, Canada (e-mail: yimin.yang@uwo.ca). }
    \thanks{Code is available at \url{https://github.com/Legendrobert/NFPF}.}
}

\markboth{IEEE TRANSACTIONS ON PATTERN ANALYSIS AND MACHINE INTELLIGENCE, VOL.X, NO.X, X~2021
}{Yuxi \MakeLowercase{\textit{et al.}}: Unsupervised Active Learning via Natural Feature Progressive Framework}

\maketitle

\begin{abstract}
The effectiveness of modern deep learning models is predicated on the availability of large-scale, human-annotated datasets, a process that is notoriously expensive and time-consuming. While Active Learning (AL) offers a strategic solution by labeling only the most informative and representative data, its iterative nature still necessitates significant human involvement. Unsupervised Active Learning (UAL) presents an alternative by shifting the annotation burden to a single, post-selection step. Unfortunately, prevailing UAL methods struggle to achieve state-of-the-art performance. These approaches typically rely on local, gradient-based scoring for sample importance estimation, which not only makes them vulnerable to ambiguous and noisy data but also hinders their capacity to select samples that adequately represent the full data distribution. Moreover, their use of shallow, one-shot linear selection falls short of a true UAL paradigm. In this paper, we propose the Natural Feature Progressive Framework (NFPF), a UAL method that revolutionizes how sample importance is measured. At its core, NFPF employs a Specific Feature Learning Machine (SFLM) to effectively quantify each sample's contribution to model performance. We further utilize the SFLM to define a powerful Reconstruction Difference metric for initial sample selection. Our comprehensive experiments show that NFPF significantly outperforms all established UAL methods and achieves performance on par with supervised AL methods on vision datasets. Detailed ablation studies and qualitative visualizations provide compelling evidence for NFPF's superior performance, enhanced robustness, and improved data distribution coverage.
\end{abstract}

\begin{IEEEkeywords}
Active learning, Unsupervised learning, Progressive framework, Autoencoder.
\end{IEEEkeywords}

\section{Introduction}
\IEEEPARstart{T}{he} success of deep learning in various domains such as computer vision, natural language processing, and medical imaging has been largely driven by the availability of large-scale, human-annotated datasets\cite{ref1}. Informative labeled data with high-quality is essential for the model evaluation. When architectural parameters remain constant, enhancing the quality of training corpora can yield substantial improvements in model performance\cite{DBLP:conf/iclr/MaharanaYB24}\cite{zheng2022coverage}. However, obtaining such labeled data is often expensive, time-consuming, and labor-intensive, especially in domains that require expert-level annotation\cite{hong2021active}. 
This data bottleneck presents a critical challenge in the development of scalable and generalizable deep learning models and large language models. In many practical scenarios, annotation budgets are limited, making it infeasible to label vast amounts of data. To address this challenge, several research directions have emerged to mitigate the dependency on large-scale labeled data, including semi-supervised learning\cite{10537231}\cite{10138923}, few-shot learning\cite{10239698}, core-set selection\cite{author2025title}\cite{zheng2025elfs} and active learning\cite{gu2020efficient}\cite{liu2021online}.Among these, \textit{active learning} (AL) offers a particularly appealing solution by aiming to select the most informative or representative samples for annotation, thereby achieving high model performance with minimal labeling cost. In addition to reducing annotation effort, AL also strives to ensure that the selected samples accelerate the convergence of downstream tasks, enabling them to reach the target performance level as quickly and efficiently as possible.

AL methods have gained increasing attention in the machine learning and computer vision communities due to its potential to reduce annotation cost while maintaining model performance. It has been successfully applied to a variety of tasks\cite{7508942}\cite{9739135}\cite{10251045}. Generally, AL methods can be categorized into two broad types: \textit{supervised} and \textit{unsupervised}. Supervised AL methods typically select informative samples based on estimated training difficulty scores. However, these methods require ground-truth labels to compute difficulty scores, which limits their applicability in fully unlabeled scenarios and undermines the goal of reducing manual annotation \cite{zong2024bidirectional}\cite{ning2022active}. In addition, they still suffer from several critical challenges, including inefficient human annotation processes, sensitivity to outliers, and the presence of noisy oracles , all of which further hinder their scalability and robustness in real-world applications\cite{li2024survey}.
\textit{Unsupervised active learning} (UAL) aims to identify informative samples without relying on labeled supervision, offering greater annotation efficiency than traditional AL methods by requiring only a one-time labeling of selected samples, without iterative training and repeated annotation across multiple rounds. Despite its increased difficulty, it offers the advantage of achieving comparable results with substantially lower labeling costs. 

\begin{figure}[t]
    \centering
    \includegraphics[width=0.7\linewidth]{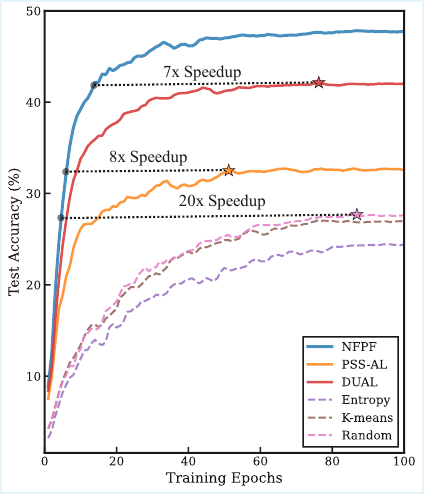}
    \caption{Fast convergence on the large-scale classification dataset CIFAR-100. The Natural Feature Progressive Framework (NFPF) requires fewer gradient steps to train the classifier compared with standard uniform data selection. PSS-AL denotes the Unsupervised Projected Sample Selector~\cite{pi2024unsupervised}, and DUAL denotes On Deep Unsupervised Active Learning~\cite{ijcai2020p364}.
}
    \label{fig:tensor1}
\end{figure}

Existing UAL approaches can be broadly categorized into two types. The first assumes that each data point can be linearly reconstructed from a selected subset\cite{zhang2011active}\cite{li2018joint}\cite{pi2024unsupervised}, which limits its applicability to non-linear data distributions. 
The second approach leverages deep learning models to project data into a latent embedding space, where discriminative feature representations are learned. Samples are then ranked based on geometric properties in the embedding space, typically under reconstruction-based constraints to ensure representativeness\cite{ma2022deep}\cite{ijcai2020p364}.  Nevertheless, these methods generally depend on computationally intensive end-to-end training frameworks with complex architectures, which incur prohibitive computational costs and cause efficiency bottlenecks. 
Furthermore, the resulting set of selected samples for labeling often exhibits class imbalance, a challenge that becomes more pronounced as the number of classes increases, thereby exacerbating the difficulty of subsequent classification tasks. 
And another key challenge of these models is their inability to effectively filter out noisy or redundant samples\cite{author2025title}\cite{mindermann2022prioritized}, which are frequently incorporated into the selected subset and hinder the improvement of model performance when training on the selected subset. 
Therefore, sample selection ought to benefit from quantifying each data point's contribution to generalization loss reduction, thereby enabling the identification of informative samples while excluding ambiguous or noisy instances based on the model's current state. Consequently, the identification of learnable samples constitutes a major open problem in UAL tasks.

In this paper, we propose a novel UAL framework, termed the Natural Feature Progressive Framework (NFPF), which leverages inter-model discrepancy loss to effectively evaluate the learnability of candidate training samples. The framework consists of two equally important components: initial subset selection and sample importance measurement. Initially, we use k-means to allocate different class cores to initialize the "weak" two layers auto-encoder named as Specific Feature Learning Machine (SFLM), and then We introduce an impactful indicator, the Reconstruction Difference (RD), to establish a fundamental approach for identifying the most challenging samples near the decision boundary. By substituting the reconstructed quality of the decision surface with a distance-difference criterion, RD precisely identifies a feature's proximity to the decision boundary. The indicator reaches its minimum value when a feature resides exactly on the boundary of its two nearest cluster centers, signifying the most ambiguous sample to differentiate. The NFPF framework takes the initial RD-selected subset as its foundation. From this subset, it progressively measures sample selection importance by employing SFLM with different neuron sizes, thereby identifying and selecting only the most informative samples. By integrating these two components, NFPF effectively captures samples that are most valuable for model training.
We explore NFPF through extensive experiments on nine datasets and evaluate the reduction in required training steps compared with classical sampling methods such as Entropy, Random, and K-means, as well as other state-of-the-art UAL approaches. CIFAR-100 is one of the dataset in our evaluation, and our NFPF achieves the target accuracy with $7\times$ to $20\times$ fewer steps than other methods, while also reaching higher accuracy when the target subset is 4000, as illustrated in Fig.~\ref{fig:tensor1}.

The main contributions of this paper are highlighted as fellows:
\begin{enumerate}

\item We introduce the Reconstruction Difference (RD) as a effective indicator for positioning sample near decision surface. With the lightweight architecture of a single layer autoencoder called Specific Feature Learning Machine (SFLM), it presumes the learned highly distinctive representation are important for discrimination. To the best of our knowledge, our work is the first to leverage model reconstruction error as a metric for quantifying sample informativeness.

\item Natural Feature Progressive framework (NFPF) is proposed to establish a novel connection between UAL and inter-model discrepancy, effectively leveraging this relationship to guide informative sample selection. It prioritizes the selection of representative and informative samples while simultaneously reducing the likelihood of selecting low-contribution or noisy instances that provide limited benefit to downstream tasks. Without any backward computation, NFPF achieves superior performance with only a single forward pass. 

\item Experimental results demonstrates that our proposed NFPF achieves outstanding performances compared with existing unsupervised active learning methods and even reaches the comparable performance with some supervised active learning algorithms. Furthermore, we perform comprehensive analyses from multiple perspectives, showing that NFPF can rapidly and reliably select highly informative samples with minimal computational cost.

\end{enumerate}
The reminder of this paper is designed as fellows: Section 2 gives a brief introduction of currently UAL and the applications of the reducible loss. Section 3 describe the steps of our proposed NFPF model in details. Section~4 presents the experimental results on the employed datasets, along with additional ablation studies and parameter sensitivity analyses. To further validate the theoretical soundness and practical effectiveness of our approach, we also conduct robustness experiments and evaluate the effectiveness of the selected subset under high implicit gradient conditions. Section~5 concludes the paper and outlines potential directions for future work.

\section{RELATED WORK}
In this section, a brief introduction of unsupervised active learning and reducible loss.
\subsection{Unsupervised Active Learning}

\begin{figure}
    \centering
    \includegraphics[width=\linewidth]{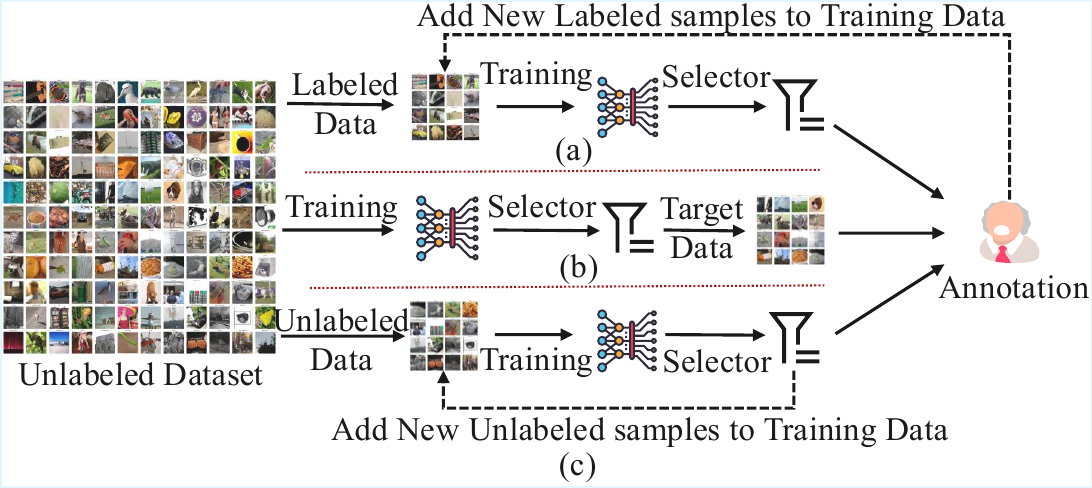}
    \caption{Active and Unlabeled Active Learning (UAL) schemes. 
        UAL aims to select an informative and representative subset without ground-truth labels, minimizing human annotation. 
        (a) Active Learning process, (b) classical UAL methods, (c) our NFPF scheme. Our proposed NFPF scheme follows an identical procedure to Active Learning, with the exception of the iterative human annotation step.}
    \label{fig:tensor2}
\end{figure}
When the annotation budget is limited, most active learning methods fail to outperform random selection, particularly those based on uncertainty sampling, which often suffer from the so-called "cold start" problem \cite{chen2022making}. Nevertheless, UAL can reduce budget dependence while remaining faithful to the distribution of unlabeled samples. Existing UAL methods can be broadly categorized into two groups. In the early stage, most approaches relied on traditional mathematical formulations. Among them, Transductive Experimental Design (TED)~\cite{10.5555/2540128.2540354} laid the foundational framework for UAL. Following this, Active Learning via Neighborhood Reconstruction (ALNR)~\cite{hu2013active} exploits neighborhood information to model the relationships between samples and construct informative subsets. Similarly, Local Structure Reconstruction (LSR) \cite{li2017active}leverages local geometric structures to guide the selection of representative data points. 
Formally, given an input dataset $\mathbf{X} \in \mathbb{R}^{d \times u}$, subset data $\mathbf{X}_S \in \mathbb{R}^{d \times m}$, where $u$ and $m$ is the number of unlabeled samples and $d$ denotes the feature dimension, our goal is to identify a representative subset $\mathbf{X}_S \subset \mathbf{X}$ that maximizes the representational coverage of $\mathbf{X}$. The method formulates the objective as:

\begin{equation}
\min_{\mathbf{S}, \mathbf{A}} \sum_{i=1}^{N} \left( ||\mathbf{x}_i - \mathbf{X}_S\mathbf{a}_i||_2^2 + \alpha ||\mathbf{A}||_{2,1} \right)
\label{eq:qe1} 
\end{equation}
where $\mathbf{X}_S \subseteq \mathbf{X}$ represents the selected subset of samples, and $\mathbf{A} = [\mathbf{a}_1, \ldots, \mathbf{a}_u]$ denotes the corresponding reconstruction coefficient matrix. 

Other approaches incorporate feature selection into convex optimization frameworks to further constrain the sample selection process named Active Learning with Feature Selection (ALFS)\cite{li2018joint}. To address the noise in low-dimensional while selecting the samples, Active Learning via Subsppace Learning (ALSL) use the low-rank structure. With the rapid advancement and powerful capabilities of deep learning, deep models have been increasingly explored to address UAL problems. One of the earliest attempts is the model proposed in on Deep Unsupervised Active Learning (DUAL)\cite{ijcai2020p364}, which was the first to utilize deep neural networks for tackling UAL tasks. Building upon this, Li~\textit{et al.} proposed several deep UAL frameworks, including~\cite{li2021deep, li2022structure}, which further advanced the development of deep learning-based UAL methods. 

Although these approaches have achieved strong performance, they primarily rely on nonlinear transformations via auto-encoders, focusing heavily on feature extraction capabilities. By imposing various constraints in the latent space, they aim to obtain more distinguishable features. However, they often overlook the relational structure between samples and the varying difficulty levels of learning individual instances. Moreover, these methods learn a selection matrix and require experts to label the entire subset at once after training. As shown in Fig.~\ref{fig:tensor2}, NFPF is aligned with the active learning paradigm, while maintaining an unsupervised process throughout the iterative selection.

While this process resembles unsupervised one-shot coreset selection, it does not strictly qualify as an unsupervised active learning task. This limitation leaves a significant gap in discovering more effective learning mechanisms.

\subsection{Reducible Loss}
Reducible Loss (RL) was first introduced in~\cite{mindermann2022prioritized} to accelerate training by prioritizing and selecting samples that are most valuable for learning within each data batch. Building on this idea, Sujit~\cite{sujit2023prioritizing} further incorporated reinforcement learning by utilizing loss-based feedback to guide sample selection. Talfan et.al employ it for online training data scheduling, where a portion of each incoming batch is chosen for model updates to enhance generalization \cite{evans2024bad}. Since the model is updated with only a single step on this reduced set, it may fail to fully utilize the information contained in the samples. More recently,~\cite{author2025title} extended this approach to labeled coreset selection in continual learning scenarios, demonstrating its applicability beyond standard batch settings. Meanwhile,this model further demonstrate that using reinforcement learning can serve as a proportional approximation of the negative implicit gradient, which aligns well with the objective of UAL—selecting informative subsets that are both highly learnable and non-noisy. Such samples typically accelerate model convergence and enable higher accuracy to be achieved within fewer training epochs. And the reinforcement learning-based strategies in \cite{mindermann2022prioritized} and \cite{tong2025coreset} exhibit methodological congruence with our NFPF and RD approach, as both leverage loss-based criteria for sample evaluation.

However, in UAL tasks, unselected samples remain inaccessible for annotation and cannot be labeled until the training process is complete, we aim to select a limited subset for iterative training.
This labeling strategy constraint prevents the direct application of prior methods that rely on iterative feedback, yet it motivates the development of new strategies tailored to UAL's unique setting. 

\section{METHODOLOGY}
\begin{figure*}[htb]
    \centering
    \subfloat[Subset $\mathbf{X}_S^0$ initialization by RD]{%
        \includegraphics[width=0.7\linewidth]{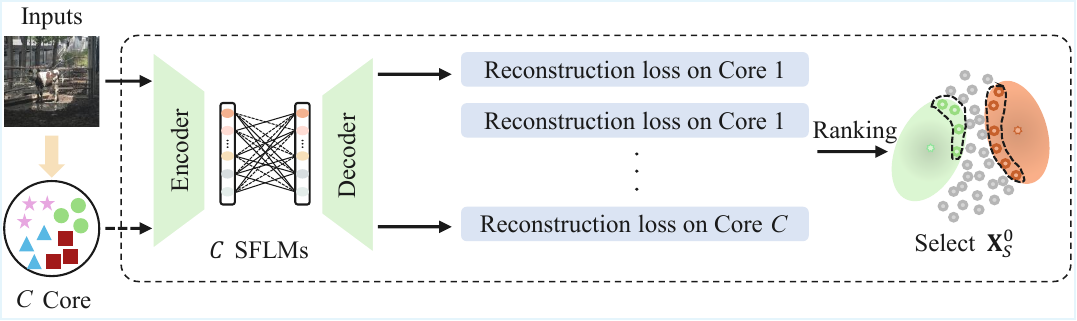}%
        \label{main_a}
    }\\ 
    \subfloat[NFPF]{%
        \includegraphics[width=0.7\linewidth]{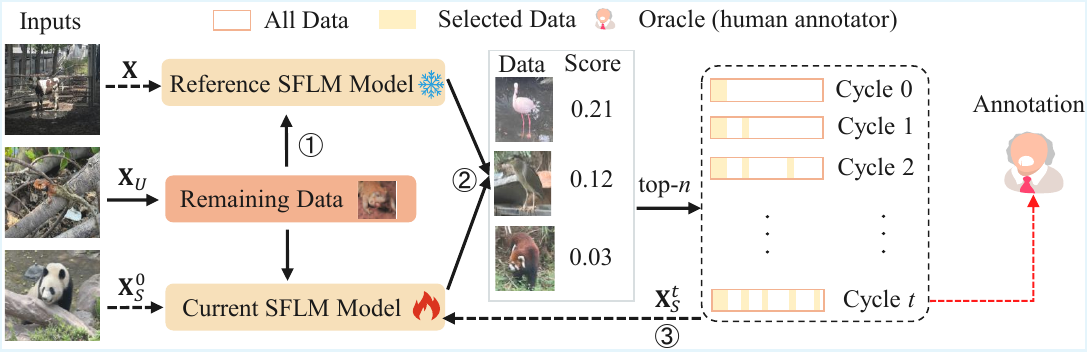}%
        \label{main_b}
    }
    \caption{The overall architecture of the proposed NFPF, showing its two main phases: a subset initialization phase and the UAL learning phase. Dashed arrows indicate the training process, while solid lines denote the direction of data flow. (a) Subset $\mathbf{X}_S^0$ initialization by Reconstruction Difference (RD) algorithm: In this phase, we first train the Specific Feature Learning Machine (SFLM) module on a core set of $C$. The trained module is then applied to all unlabeled data to compute their RD scores, enabling the ranking and selection of the initial subset. (b) NFPF scheme: The reference SFLM model is first trained on all data initially without label. The current SFLM model is trained on $\mathbf{X}_S^0$ from (a). In each learning cycle,  \scalebox{1.3}{\raisebox{-0.24ex}{\ding{192}}} calculate reconstruction loss for both reference and current model. \scalebox{1.3}{\raisebox{-0.24ex}{\ding{193}}} rank the score and select top-$n$ samples into the candidate pool $\mathbf{X}_S^t$. \scalebox{1.3}{\raisebox{-0.24ex}{\ding{194}}} train the current model for next cycle. This process is iterated $t$ times until the subset reaches the target size $m$. The accumulated selected samples are subsequently forwarded to the oracle for annotation and used for downstream classification tasks.}
    \label{fig:framework}\end{figure*}

This section introduces the cyclic framework and learning strategies of our proposed Natural Feature Progressive Framework (NFPF) for optimal data subset selection, which schematic diagram is shown in Fig.~\ref{fig:framework}. As previously discussed, certain hard samples tend to lie closer to the decision boundary and thus provide more informative signals\cite{margatina2021active}. However, this may introduce the risk of long-tailed data distribution. Recent studies have also highlighted that data coverage is a crucial factor in determining sample informativeness \cite{zheng2022coverage}\cite{maharana2023d2}. Motivated by these observations, we aim to integrate both advantages in a unified manner to enable the rapid selection of more effective samples.

We first define the notations, as illustrated in Table~\ref{tab:notations}. 
\begin{table}

\caption{\centering Notations of NFPF}

\label{tab:notations}

\centering

\footnotesize

\begin{tabular}{llc}

\hline

{Notations} & {Description} \\

\hline

{General Notations}  \\

$d$ & Dimension of features  \\

$t$ & Cycle number \\

$n$ & The number of samples acquired in $t$-th cycle \\

$\theta$ & Trained SFLM model \\

$G_i$ & The learnability score of the $i$-th sample  \\

$\mathbf{X}$ & $\mathbb{R}^{d \times u}$, unlabeled data \\

$\mathbf{X}_S$ & $\mathbb{R}^{d \times m}$, target subset\\

$\mathbf{X}_U$ & $\mathbb{R}^{d \times l}$, remaining unselected samples\\
[0.15em]

$\mathbf{X}_S^0$ & $\mathbb{R}^{d \times k}$, initial seed set\\
[0.12em]

$\mathbf{\Phi}$ & Correlation Matrix\\[0.3em]

\hline

{SFLM} & \\

$(a_h, b_h)$ & Input weight and bias in encoder \\

$\beta$ & Output weight in decoder \\

$f$ & Invertible Activation function \\

$f(aX+b)$ & Output of encoder\\

$H$ & Number of hidden neurons \\

\hline

\end{tabular}

\end{table}  
A typical Unsupervised Active Learning (UAL) scenario involves an informative data subset $\mathbf{X}_S \subset \mathbf{X}$, where $\mathbf{X}$ denotes the entire dataset. The goal of UAL is to learn an effective model $\theta$ without any labeling budget and to provide the selected samples for annotation only after the training process is completed. However, in our NFPF framework, we adopt a learning paradigm consistent with active learning. Let $\mathbf{X}_U\in \mathbb{R}^{d \times l}$ denote the set of remaining unselected samples, and $\mathbf{X}_S^0 \in \mathbb{R}^{d \times k}$ the set of the first selected samples. At the $t$-th cycle, the NFPF criterion is applied to select the top-$n$ most informative samples from the remaining unselected samples $\mathbf{X}_U^t$. 
The subset pool is then updated as
$\mathbf{X}_S^{t+1} = \mathbf{X}_S^t \cup \text{top-}n(\mathbf{X}_U^t)$
and the updated $\mathbf{X}_S^{t+1}$ is subsequently used to train a new model $\theta_{X_S}^{t+1}$.
This procedure repeats until the total number of selected samples reaches $m$, after which all selected samples are submitted for annotation.

\subsection{Natural
Feature Progressive Framework}
\label{section3.1}
To derive the gradient with respect to A, we first transform Eq.~\ref{eq:qe1} into the following form:
\begin{equation}
\label{Eq.2}
\nabla_{\mathbf{A}} \mathcal{L} = 2\mathbf{X}^{\top}(\mathbf{X}\mathbf{A} - \mathbf{X}) + \alpha \mathbf{D}^{-1} \mathbf{A}
\end{equation}
where $\mathbf{D}$ is a diagonal matrix with diagonal elements $D_{ij} = \|\mathbf{a}_j\|_2 = \sqrt{\sum_{i=1}^{N} A_{ij}^2}$. A larger reconstruction error indicates that the sample is difficult to represent using existing data. This forces the model to make larger adjustments during gradient updates, making the sample's influence on the overall optimization more significant. In this sense, the selection strategy can be viewed as a reformulation of a gradient-based scoring mechanism, where each candidate sample is evaluated based on its potential to improve the original learning objective. Thus, an intuitive prioritization strategy favors hard-to-learn examples, whereas easy ones produce only small gradients. The core issue is that both noisy and highly informative samples can yield large gradients. This leads us to question: can we directly capture the reduction of the outer objective in Eq. \ref{Eq.2} as the selection criterion?

We define the gain of selecting a new sample $\mathbf{u}_i$ from $\mathbf{X}_U$ based on the improvement in its log-likelihood under the model trained with the updated subset. We define two SFLM models: $\theta_{X_S}$ is the current model with the initial $h$ hidden neurons, which is trained on the subset $\mathbf{X}_S$ with a small number of hidden nodes. For comparison, $\theta_X$ serves as the reference model with $10*h$ hidden neurons, trained on the entire dataset with a larger number of hidden nodes. Subsequently, the learnable prioritization of samples can be expressed in a compact form $G^{hard}=\ell(\mathbf{u}_i;\theta_{X_S})$. Conversely, giving preference to samples that are readily solved by a well-trained model can effe ctively eliminate noisy instances present in large datasets $G^{easy}=-\ell(\mathbf{u}_i;\theta_X)$. These definitions suggest that the same model learns features with different distributions—local or global—depending on the scale and quality of the training dataset. As a result, a new sample processed by different models undergoes generalization according to each model’s learned capabilities. This underlying concept is consistently employed in multimodal learning for assessing the quality of samples in pre-trained models~\cite{schuhmann2022laion, hessel2021clipscore}.  

\begin{algorithm}[t!]
\caption{The Proposed Algorithm of NFPF}
\label{alg:nfpf}
\hspace*{\algorithmicindent} \textbf{Initialization:} Given $\mathbf{X}$, $\mathbf{X}_S^0$, $\mathbf{X}_U$, $m$, $n$, ${C}$, ${t}$, $G$ \\

\begin{algorithmic}[1]
\State{Train reference model \( {\theta}_{\mathbf{X}} \) by SFLM on $\mathbf{X}$}
\State{K-Means cluster centers $C$ to train $C$ SFLMs}
\For{$c \gets 1$ to ${C}$}
    \State{Compute RDs by (\ref{RD})}
\EndFor
\State{Sort the $\mathbf{X}$ in ascending order and get the $\mathbf{X}_S^0$}
\State{Train current model \( {\theta}_{\mathbf{X}_S}^0 \) by SFLM on $\mathbf{X}_S^0$}
\State{Define $\mathrm{T} = (m-k)/n$}
\For{$t \gets 1$ to $\mathrm{T}$}
\State{Compute the learnability score $G$ by (\ref{score})}
\State{Sort the unlabeled dataset $\mathbf{X}_U^t$ in ascending $G$ order}
\State{$\mathbf{X}_S^t\gets$ $\mathbf{X}_S^{t-1}\cup$ top-$n(\mathbf{X}_U^t)$ }
\State {Train current model \( {\theta}_{\mathbf{X}_S}^t \) by SFLM on $\mathbf{X}_S^t$}
\EndFor
\end{algorithmic}

\end{algorithm}

We then define a selective score to measure a sample's learnability:
\begin{equation}
\label{NFPF Score}
G_i=G^{hard}+G^{easy}=\ell(\mathbf{u}_i;\theta_{X_S})-\ell(\mathbf{u}_i;\theta_X)
\end{equation}
where $\ell(\mathbf{u}_i;\theta_{X_S})$ and $\ell(\mathbf{u}_i;\theta_X)$ are the losses of sample $\mathbf{u}_i$ on the models trained on the subset $\theta_{X_S}$ and the full dataset $\theta_X$, respectively.
We use the reconstruction correlation coefficient, denoted by $\phi()$, to represent the sample selection score, which is detailed in Section~\ref{section3.2}. Eq.~\ref{NFPF Score} then can be equivalently expressed as:
\begin{equation}
\label{score}
G_i= \phi(\mathbf{u}_i;\theta_{X_S})- \phi(\mathbf{u}_i;\theta_X)
\end{equation}
The core idea behind our sample selection strategy is that a sample $x_i$ is considered \textit{representative} if its reconstruction loss under $\theta_{\mathbf{X}}$ is low, and \textit{informative} if its loss under $\theta_{\mathbf{S}}$ is high. This suggests that $x_i$ brings novel information to the current subset $\mathbf{X}_S$. Intuitively, adding such a sample to $\mathbf{X}_S$ would lead to the largest reduction in reconstruction loss, which is approximately equivalent to maximizing the gradient signal during training.

The schematic diagram of the NFPF is shown in Fig.~\ref{fig:framework}, and the detailed processes are as follows:  
\begin{enumerate}
    \item Initalize $\mathbf{X}_S^0$ by RD, as depicted in Fig.\ref{fig:framework}\subref{main_a}. 
    \item Train a reference model $\theta_{X}$ on unlabeled $X$.
    \item Train a current model $\theta_{X_S}$ on $\mathbf{X}_S^0$.
    \item Gain the remaining samples $\mathbf{X}_U^0$=$\mathbf{X}-\mathbf{X}_S^0$.
    \item Compute the sample learnability score $G_i$ while a new sample $\mathbf{u}_i$ is added and rank the samples $\mathbf{X}_U^0$ according to the resulting values.
    The top $n$ samples $\mathbf{X}_U^0$ are then selected as new candidates and added to $\mathbf{X}_S^1$. 
    \item Training current model on $\mathbf{X}_S^1$.
    \item Repeat Step 4,5 and 6 until meeting labeling budgets.
\end{enumerate}
In each cycle, we sample from the remaining-pool $\mathbf{X}_U$. After each selection step, the remaining samples are returned to the candidate pool for the next iteration. Since all samples are repeatedly selected and reintroduced into the pool, the procedure not only covers the full data distribution but also ensures novelty. The initial subset $\mathbf{X}_S^0$ used to train the current model plays a crucial role. In the following, 
Section~\ref{section3.3} elaborates on the procedure for generating the initial subset, while Section~\ref{section3.2} provides a detailed description of the SFLM formulation, as well as the methodology for constructing the current and reference models.

\subsection{Subset Initialization via Reconstruction Difference}
\label{section3.3}
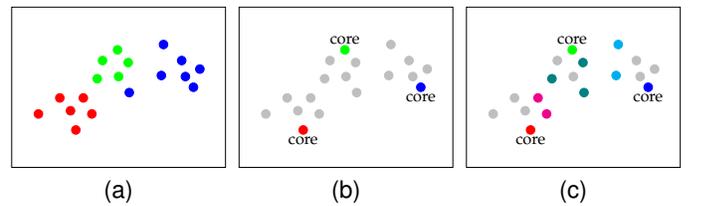
\begin{figure}[!b]
    \centering
    \subfloat[\label{fig:GroundTruth}]{
        \resizebox{0.32\linewidth}{!}{%
            \begin{tikzpicture}
                \draw [black,ultra thin] (0,0) rectangle (4,3); 
                \foreach \x/\y in {1.2/0.7,1.5/1,0.5/1,1.34/1.3,1.1/1.06,0.9/1.3}
                    \draw [red,fill] (\x,\y) circle [radius=0.08];
                \foreach \x/\y in {3.4/1.5,3.25/1.7,3.52/1.84,2.2/1.4,3.18/2.0,2.8/1.72,2.84/2.3}
                    \draw [blue,fill] (\x,\y) circle [radius=0.08];
                \foreach \x/\y in {2.0/1.7,1.7/2.0,2.18/1.962,1.6/1.66,1.98/2.2}
                    \draw [green,fill] (\x,\y) circle [radius=0.08];
            \end{tikzpicture}%
        }
    }%
    \subfloat[\label{fig:Phase1init}]{
        \resizebox{0.32\linewidth}{!}{%
            \begin{tikzpicture}[every node/.style={font=\small}]
                \draw [black,ultra thin] (0,0) rectangle (4,3); 
                \draw [red,fill] (1.2,0.7) circle [radius=0.08] node [black,below=0.1] {core}; 
                \foreach \x/\y in {1.5/1,0.5/1,1.34/1.3,1.1/1.06,0.9/1.3}
                    \draw [lightgray,fill] (\x,\y) circle [radius=0.08];
                \draw [blue,fill] (3.4,1.5) circle [radius=0.08] node [black,below=0.1] {core};
                \foreach \x/\y in {3.25/1.7,3.52/1.84,2.2/1.4,3.18/2.0,2.8/1.72,2.84/2.3}
                    \draw [lightgray,fill] (\x,\y) circle [radius=0.08];
                \draw [green,fill] (1.98,2.2) circle [radius=0.08] node [black,above=0.1] {core}; 
                \foreach \x/\y in {2.0/1.7,1.7/2.0,2.18/1.962,1.6/1.66}
                    \draw [lightgray,fill] (\x,\y) circle [radius=0.08];
            \end{tikzpicture}%
        }
    }%
    \subfloat[\label{fig:Phase1result}]{
        \resizebox{0.32\linewidth}{!}{%
            \begin{tikzpicture}[every node/.style={font=\small}]
                \draw [black,ultra thin] (0,0) rectangle (4,3); 
                \draw [red,fill] (1.2,0.7) circle [radius=0.08] node [black,below=0.1] {core}; 
                \draw [blue,fill] (3.4,1.5) circle [radius=0.08] node [black,below=0.1] {core}; 
                \draw [green,fill] (1.98,2.2) circle [radius=0.08] node [black,above=0.1] {core}; 
                \foreach \x/\y in {1.34/1.3,1.5/1} \draw [magenta,fill] (\x,\y) circle [radius=0.08]; 
                \foreach \x/\y in {2.2/1.4,2.18/1.962,1.6/1.66} \draw [teal,fill] (\x,\y) circle [radius=0.08]; 
                \foreach \x/\y in {2.8/1.72,2.84/2.3} \draw [cyan,fill] (\x,\y) circle [radius=0.08]; 
                \foreach \x/\y in {0.5/1,1.1/1.06,0.9/1.3,3.25/1.7,3.52/1.84,3.18/2.0,2.0/1.7,1.7/2.0}
                    \draw [lightgray,fill] (\x,\y) circle [radius=0.08];
            \end{tikzpicture}%
        }
    }
    \caption{Procedure of subset initialization: (a) Ground Truth; (b) 
    Three core categories (red, blue, green) are selected to train the SFLM model; (c) $\mathbf{X}_S^0$ results ($k=10$).  Some samples that lie close to two cores are highlighted in magenta, cyan, or teal, while gray points denote remaining data.}
    \label{fig:phase1}
\end{figure}

In this section, we propose the Reconstruction Difference (RD) algorithm, designed to initialize $\mathbf{X}_S^0$ through the selection of boundary-near samples. The schematic diagram of the RD is shown in Fig.~\ref{fig:phase1}.

As suggested in~\cite{maharana2023d2,sorscher2022beyond}, we assume that the number of classes $C$ is known a priori. This assumption reflects realistic deployment scenarios where domain knowledge provides structural constraints, consistent with unsupervised learning frameworks that leverage predefined semantic category cardinality. Under this setting, we initialize our method by applying K-means\cite{mcqueen1967some} clustering to obtain $C$ cluster centers as starting points.  We begin by training $C$ SFLMs, with parameters denoted by $\theta_1, \dots, \theta_C$. Each model is trained on the c-core data, and its training performance is summarized by a scalar score $\alpha_c$. These scores are collected into a training score vector $\bm{\alpha} = [\alpha_1, \dots, \alpha_C]$. we utilize the remaining unselected samples $\mathbf{X}_U$, comprising $t$ test samples. For each sample $\mathbf{u}_i \in \mathbf{X}_U$ and each $\text{SFLM}_c$, we compute its reconstruction correlation coefficient $\phi$. These scores are organized into an score matrix $\bm{\Phi}$, where each element $\Phi_{ic}$ represents the reconstruction correlation coefficient of subnetwork $C$ on test sample $\mathbf{u}_i$. The matrix can be formally expressed as:
\begin{equation}
   \bm{\Phi}_{iC} = \phi(\mathbf{u}_i;\theta_C(\mathbf{u}_i))
\end{equation}
And the distance vector for each $\theta_C$ to quantify the difference between c-core data and other data performance. For each $\theta_C$, we compute a distance vector $\mathbf{d}_C$, whose elements measure the absolute difference between $\alpha_C$ and its performance on each test sample:
\begin{equation}
    \mathbf{d}_i = \begin{bmatrix}
|\phi_{1C} - \alpha_C| \\
|\phi_{2C} - \alpha_C| \\
\vdots \\
|\phi_{iC} - \alpha_C|
\end{bmatrix}
\end{equation}
This vector $\mathbf{d}_i$ provides a fine-grained measure of $C$'s SFLM generalization performance across the entire test set. Although this approach does not account for certain scenarios, such as non-convex decision boundaries or overlapping clusters, it can still capture samples located near the boundary hyperplane. In such a simplified scenario, boundary-near samples can be further characterized as those whose distances to their nearest and second-nearest cores are approximately equal.
Formally, the boundary sample set can be defined as RD indicator:
\begin{equation}
S = \left\{ \mathbf{u}_i \in \mathbf{U} \;\middle|\;
\min_{c_j \ne c^*} \left| \Phi_{tc_j} - \alpha_{c_j} \right|
\;\approx\;
\left| \Phi_{tc^*} - \alpha_{c^*} \right|
\right\},
\end{equation}
where $c^*$ denotes the nearest core of sample $\mathbf{u}_i$ and $S$ denotes the set of samples near the decision boundary. 
To make the definition more precise, let $c_1$ and $c_2$ be the nearest and the second-nearest centers of $\mathbf{u}_i$, respectively. Then the boundary set can be expressed as
\begin{equation}
\label{RD}
S = \Big\{ \mathbf{u}_i \;\Big|\; 
RD=\big|\, d_{ic_1} - d_{ic_2} \,\big| \leq \varepsilon
\Big\},
\end{equation}
where $d_{ic} = \lvert \Phi_{i c} - \alpha_{c} \rvert$. Here, $\big|\, d_{ic_1} - d_{ic_2} \,\big|$ denotes our proposed Robustness-Diversity (RD) indicator, and $\varepsilon$ serves as a tolerance threshold for assessing the proximity of distances. In the ideal scenario, $\varepsilon \rightarrow 0$, signifying that the samples lie precisely on the decision boundary. The initial subset $\mathbf{X}_S^0$ is then constructed by selecting samples from $\mathbf{X}$ in a descending order of their RD values, prioritizing those with the highest RD scores, which correspond to the most challenging-to-discriminate samples.

\subsection{Specific Feature Learning Machine}\label{section3.2}
\begin{figure}   
  \centering
  \includegraphics[width=0.8\linewidth]{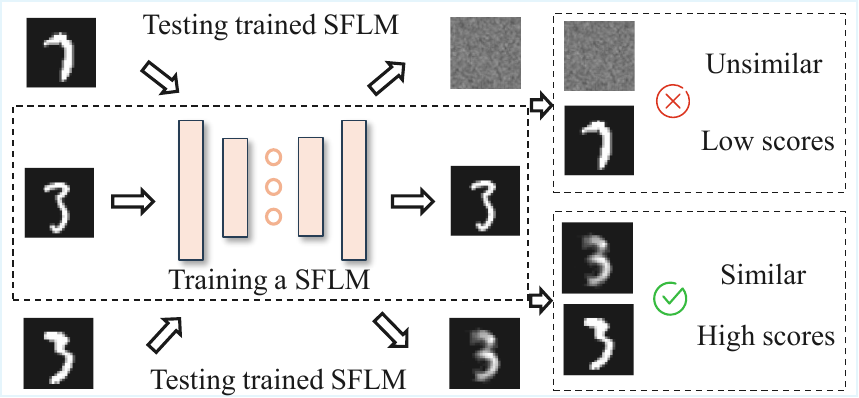}
  \caption{Idea of the cluster growing with reconstructed samples; The key is to design a fast but "weak" autoencoder}
  \label{fig:moti1}
\end{figure}

Our approach is inspired by the work of \cite{YangPLM}, which utilizes single-hidden-layer feedforward networks (SLFNs) to progressively approximate hybrid nonlinear/linear systems. This method classifies training data into continuous clusters, effectively grouping similar samples. Building upon this, we introduce a novel \textbf{indirect similarity metric} to quantify the relationship between original and reconstructed data points. We achieve this by reformulating our objective to leverage the reconstruction error as a measure of inter-sample distance. For this purpose, we employ the SFLM, a two-layer network with a non-iterative learning strategy, following the work in \cite{yang2015extreme}. A key aspect of our approach is its simplicity: we use a single subnetwork node composed of just $h$ hidden units. We deliberately forgo the full universal approximation capabilities of more complex models to demonstrate that a streamlined architecture can still achieve effective clustering performance, as illustrated in Fig.~\ref{fig:moti1}.

\subsubsection {SFLM Architecture} 
The notation $\left\{ \left( \mathbf{a}_h, b_h, \beta_h \right) \right\}_{h=1}^{H}$ represents the parameters partition of SFLM, $H$ denotes the number of hidden nodes, $\mathbf{a}_h$ the input weight vector, $b_h$ the bias, and $\beta_h$ the output weight.
The output weight matrix $\beta$ can be computed as follows:
\begin{equation}
    \begin{array}{l}
      \beta=\mathbf{H}^{\dagger}\mathbf{X}\\
    \end{array}
\end{equation}
for which $\mathbf{H}^{\dagger}$ is called a Moore Penrose Pseudo-inverse matrix and $\mathbf{H}$ denotes the output of hidden layer. Such a pseudoinverse matrix are calculated as follows
\begin{equation}
    \left \{
        \begin{array}{ll}
          \mathbf{H}^{\dagger}=\mathbf{H}^{\mathbf{T}} \cdot {(\frac{\mathbf{C}}{\mathbf{I}} + \mathbf{H} \mathbf{H}^\mathbf{T} )}^{-1}, & \mathbf{H}^{T} \mathbf{H} \mbox{ is singular} \\
          \mathbf{H}^{\dagger}= {(\frac{\mathbf{C}}{\mathbf{I}} + \mathbf{H} \mathbf{H}^\mathbf{T} )}^{-1} \cdot \mathbf{H}^{\mathbf{T}} , & \mathbf{H}^{T} \mathbf{H} \mbox{ is non-singular}
        \end{array}
    \right.
\end{equation}
After calculating $\beta$, the input weights and bias are also updated based on our previous method \cite{yang2015extreme}.
\begin{equation}
\begin{split}
&a=g^{-1}(u(\textbf{e}))\cdot \mathbf{X}^{T}(\frac{C}{\mathbf{I}}+\mathbf{X}\mathbf{X}^{T})^{\dagger}\\
&\beta=\mathbf{H}^{\dagger}\mathbf{X}\\
   & f^{-1}(\cdot) \left \{
        \begin{array}{ll}
          =\arcsin(\cdot),\, \text{if} \,\,f(\cdot)=sin(\cdot) \\
          = -\log(\frac{1}{(\cdot)}-1), \, \text{if} \,\,h(\cdot)=1/(1+e^{-(\cdot)}) \\
        \end{array}
    \right.
\end{split}
\end{equation}
Note that the positive parameter $\mathbf{C}=[2^{-10},2^{10}]$ is used as a regularization term to improve the stability. 
Therefore, the distances between the remaining samples  and the sample selected $n$ to be trained can be measured using:
\begin{equation} \label{mindis}
  \left.
    \begin{array}{l}
      \displaystyle\min_{\theta}E(\theta)=\displaystyle\sum_{i=1}^{k} \displaystyle\sum_{j=1}^{m} D(\mathbf{x}_{ij}, f(\mathbf{x}_{ij},\mathbf{x}_i,\theta))\\
    \end{array}
  \right.
\end{equation}
where $D( \cdot\; ;\theta)$ indicates pairwise dissimilarities in which $\theta$ represents the trained SFLM model, a simplified notation for the set of learned parameters $\left\{ \left( \mathbf{a}_h, b_h, \beta_h \right) \right\}_{h=1}^{H}$. and $f(x_{ij},x_i,\theta)$ signifies reconstructed members via the SFLM learned by $x_i$. A statistical similarity metric, correlation, is employed to quantify the reconstruction distance of samples, reflecting the performance of $D(\cdot \, ; \theta)$, and subsequently serving as a surrogate for their actual distance in the feature space.
\begin{equation} 
\label{corr}
    \phi_{ij} = corr (x_{ij}, \hat{\mathbf{x}}_{ij}) = \dfrac{ cov (\mathbf{x}_{ij}, \hat{\mathbf{x}}_{ij})} {\sigma_{\mathbf{x}_{ij}} \cdot \sigma_{\hat{\mathbf{x}}_{ij}}}
\end{equation}
where $\mathrm{cov}(\cdot)$ denotes the covariance, $\sigma_{\mathbf{x}_{ij}}$ and $\sigma_{\hat{\mathbf{x}}_{ij}}$ represent the standard deviations of the two variables, 
and $\phi_{ij} \in [0,1]$, with larger values indicating closer proximity (higher similarity) 
and smaller values indicating greater distance (lower similarity) between $\mathbf{x}_{ij}$ and $\hat{\mathbf{x}}_{ij}$. 
By using SFLM, hidden neurons retain specialized features provided by ${x_i}$ which are most characteristic to present the hidden pattern. From our prospective, these features are highly shareable within the class. That is supposing the input ${\mathbf{x}_i}$ is used to train the SFLM, then an unlabeled $\mathbf{x}_{ij}$ is feed-forward into that SFLM. After encoding, the squashed hidden vector is projected back by applying the trained decoding mapping $p(\mathbf{x}_i|h_i)$. $\mathbf{x}_{ij}$ can only be well-reconstructed when the used decoding function approximates to expected, i.e. a desired decoding distribution $p(\mathbf{x}_{ij}|h_{ij}) \approx p(\mathbf{x}_i|h_i)$. 
The resulting SFLM, derived from this design philosophy, serves as a principled and theoretically grounded component, forming the cornerstone of our overall framework.

\subsubsection {Train Current and Reference Models with SFLM} 
Informax approach has been a long held belief in field of representation learning, for which its underlying ideology is to retain exhaustive information through the parameterized mapping from input $\mathbf{X}$ to a latent representation ${H}$. With the goal of generalization capability, most auto-encoders are dedicated to model generative non-linear function, for which unseen instances are capable to be compressed into and decompressed from the latent space.

The Reference Model, denoted as $\theta_X$, is an SFLM trained directly on the entire dataset $\mathbf{X}$. To enable the model to fully learn the complete sample distribution, we set its hidden layer size, $H$, to a large value, allowing it to serve as a robust baseline for measuring the reconstruction score of all samples. In contrast, the Current Model, denoted as $\theta_{X_S}$, is a more generalized version with a smaller hidden layer size. It is trained on the cumulative subset from each cycle, providing a local feature perspective.
Given this design, a sample $\mathbf{u}_i$ can be considered representative if its reconstruction loss, $\ell(\mathbf{u}_i;\theta_X)$, is low. Conversely, a sample $\mathbf{u}_i$ is deemed informative if its reconstruction loss, $\ell(\mathbf{u}_i;\theta_{X_S})$, is high, as this indicates it contains new information not yet captured by $X_S$.

\section{EXPERIMENTS} 
\label{section4}
In this section, we conduct extensive experiments for different labeling budgets to evaluate the effectiveness of the proposed Natural Feature Progressive Framework (NFPF) in comparison with state-of-the-art (sota) UAL methods and traditional methods. In Section \ref{UCI}, we conduct experiments on widely used public datasets and compare our method against state-of-the-art baselines.  
In Section \ref{OBJECT}, we further evaluate our approach on image datasets to validate its effectiveness, and, for a more comprehensive comparison, we additionally include two state-of-the-art supervised active learning methods.  
In Section \ref{visual}, we provide visualizations to offer a more vivid and intuitive illustration of our results.  

\subsection{Experimental Setting} 
\label{configuration}
 \textit{Datasets details}: We conduct experiments on nine publicly available datasets spanning a variety of domains to comprehensively evaluate the effectiveness and generalizability of NFPF. These datasets encompass both structured and unstructured data types and span a variety of tasks. sEMG captures hand movements; and the Gas Sensor Array Drift (GSAD) dataset simulates sensor readings under temporal drift conditions, Epileptic Seizure Recognition (ESR) are biology time-series.    Plant Species Leaves and Letter Recognition are image-based datasets focused on fine-grained classification. The above five datasets are obtained from the UCI Machine Learning Repository\footnote{All six datasets are sourced from the University of California, Irvine (UCI) Machine Learning Repository: .\url{https://archive.ics.uci.edu/ml/datasets.php}}. In the vision domain, we include three real world image classification benchmarks: CIFAR-10\cite{krizhevsky2009learning}, CIFAR-100\cite{krizhevsky2009learning}, and Tiny-ImageNet\cite{deng2009imagenet}, which vary in complexity and visual granularity. A summary of the datasets is provided in Table~\ref{tab:wide_table}.
\begin{table*}[tbp]
\caption{\centering Summary of Datasets and Hyperparameter Settings Used in Our Experiments. $k$ and $n$ are instance \% of target subset size $m$.}
\label{tab:wide_table}
\centering
\footnotesize
\begin{tabular}{lcccccccc}
\hline
Dataset & Size & Dimension & Class & Source & $k$ (\% of $m$) & $n$ (\% of $m$) & Current model $H$ \\
\hline
sEMG                    & 3190   & 60      & 2   & UCI     & 10  & 10  & 100 \\
Plant Species Leaves    & 1600   & 64      & 100 & UCI     & 40  & 10  & 150 \\
Waveform                & 5000   & 40      & 3   & UCI     & 30  & 16  & 100 \\
ESR                     & 11500  & 178     & 5   & UCI     & 40  & 12  & 100 \\
GSAD                    & 13910  & 128     & 6   & UCI     & 40  & 5   & 150 \\
Letter Recognition      & 20000  & 16      & 10  & UCI     & 30  & 10  & 150 \\
CIFAR-10                & 60000  & 256     & 10  & Object  & 50  & 1   & 200 \\
CIFAR-100               & 60000  & 256     & 100 & Object  & 50  & 3   & 200 \\
TinyImageNet            & 100,000 & 12288   & 200 & Object  & 60  & 5   & 200 \\
\hline  
\end{tabular}
\end{table*}

\textit{Baseline}: To conduct a comprehensive comparison and validate the performance of the active learning tasks, the NFPF is compared against several sota UAL approaches, along with a few classic baselines commonly used in UAL tasks:
\begin{enumerate}[label=\alph*),leftmargin=3em,nosep,topsep=0pt, partopsep=0pt, parsep=0pt, itemsep=0pt]
    \item K-means clustering: uses the full budget to cluster into k centers and selects samples by distance.
    \item Deterministic Column Sampling (DCS) \cite{papailiopoulos2014provable}: approximates a matrix by deterministically selecting columns with the highest leverage scores.
    \item ALNR \cite{hu2013active}: selects informative samples by incorporating local geometric structure into neighborhood-based linear reconstruction.
    \item Manifold Adaptive Experimental Design (MAED) \cite{cai2011manifold}: selects samples by incorporating manifold structure into a kernel space via graph Laplacian.
    \item DUAL \cite{ijcai2020p364}: addresses unsupervised active learning sample selection by learning nonlinear features with a deep encoder-decoder.
    \item Active learning model via learnable graphs (ALLG) \cite{ma2022deep}: uses learnable graph structures to enhance representation and sample selection.
    \item Unsupervised projected sample selector (PSS-AL) \cite{pi2024unsupervised}: optimizes sample selection via orthogonal projections, balancing diversity and representativeness to enhance performance.
\end{enumerate}
These methods are evaluated on the UCI benchmark datasets. 

For the CIFAR-10, CIFAR-100, and Tiny-ImageNet datasets, DUAL and PSS-AL are selected as representative UAL baselines, given the absence of publicly available implementations for several other approaches. To further substantiate the capability of the proposed unsupervised NFPF in identifying highly informative samples for expert annotation within a practical active learning setting, we also benchmark against two cutting-edge supervised active learning algorithms:
\begin{enumerate}[label=\alph*),nosep,topsep=0pt,leftmargin=3em, partopsep=0pt, parsep=0pt, itemsep=0pt]
    \item Random sampling: simplest baseline.
    \item Noise~\cite{li2024deep}: which estimates uncertainty via noise stability by measuring output changes under parameter perturbations.
    \item Balancing active learning (BAL)~\cite{li2023bal}:a framework using self-supervised features and cluster distance difference to balance diversity and uncertainty for effective active learning.
    \item Effective label-Free coreset selection (ELFS)~\cite{zheng2025elfs}: a label-free coreset selection method leveraging deep clustering and double-end pruning to improve sample selection without ground truth labels.
\end{enumerate}

\textit{Experimental setup and implementation}: 
The first six datasets are partitioned into candidate and testing sets using a 50\%-50\% split. Datasets obtained from the UCI and Network repositories are preprocessed by their respective providers and reasonably divided into training and testing subsets. For CIFAR-10, CIFAR-100, and Tiny-ImageNet, we adhere to the standard data splits to define the candidate and test sets. To ensure consistent evaluation, the same candidate set is used across all UAL methods for selecting the top-$n$ most informative samples. The parameters setting is shown in Table~\ref{tab:wide_table}.
Following the settings in~\cite{ma2022deep, li2021deep, li2022structure}, we employ a linear SVM for the UCI datasets. For the object datasets, ResNet-18~\cite{he2016deep} is adopted as the backbone in experiments on CIFAR-10, CIFAR-100, and Tiny-ImageNet. All experiments are repeated ten times, and the results are reported as the mean classification accuracy with standard deviation. All implementations are conducted on an NVIDIA RTX 4080 GPU (12 GB) with 32 GB of system memory.

\subsection{Experiment with UCI Datasets} 
\label{UCI}

\begin{table*}
\centering
\caption{\centering Comparison of different methods on multiple datasets.}
\label{tab:UCI}
\setlength{\tabcolsep}{6pt}
\renewcommand{\arraystretch}{1.2}
\begin{tabular}{c|c|cccccccc}
\toprule
\multirow{2}{*}{Dataset} & \multirow{2}{*}{Target size $m$} & \multicolumn{8}{c}{Method} \\
 & & NFPF & ALLG & DUAL & PSS-AL & ALNR & MAED & K-Means & DCS \\
\midrule
\multirow{5}{*}{sEMG}
 & 140 & \cellcolor{gray!20}32.03 & 30.89 & 28.25 & 26.00 & 20.97 & 23.20 & 24.85 & 27.21 \\
 & 280 & \cellcolor{gray!20}34.88 & 34.76 & 27.87 & 31.89 & 23.28 & 28.64 & 23.44 & 27.14 \\
 & 420 & \cellcolor{gray!20}36.02 & 35.89 & 32.13 & 34.22 & 23.42 & 21.32 & 23.72 & 26.82 \\
 & 560 & \cellcolor{gray!20}36.42 & 34.82 & 34.27 & 35.78 & 25.06 & 27.82 & 24.92 & 27.58 \\
 & 700 & \cellcolor{gray!20}36.92 & 36.52 & 34.24 & 35.28 & 23.62 & 27.15 & 24.01 & 28.03 \\
\midrule
\multirow{6}{*}{Plant species leaves}
 & 100 & \cellcolor{gray!20}42.27 & 42.11 & 39.46 & 28.75 & 37.88 & 36.29 & 24.55 & 33.19 \\
 & 200 & \cellcolor{gray!20}59.28 & 57.65 & 54.81 & 38.62 & 52.44 & 51.76 & 39.53 & 47.90 \\
 & 400 & \cellcolor{gray!20}67.88 & 67.73 & 64.38 & 50.00 & 62.80 & 61.32 & 50.91 & 57.14 \\
 & 500 & \cellcolor{gray!20}71.27 & 70.88 & 67.45 & 54.75 & 65.21 & 64.19 & 54.60 & 60.25 \\
 & 700 & \cellcolor{gray!20}73.46 & 72.39 & 70.36 & 60.12 & 68.37 & 67.31 & 59.34 & 63.45 \\
\midrule
\multirow{5}{*}{Waveform}
 & 50  & \cellcolor{gray!20}81.22 & 81.12 & 78.91 & 76.74 & 72.26 & 78.48 & 76.81 & 76.32 \\
 & 150 & \cellcolor{gray!20}83.32 & 83.22 & 81.26 & 82.72 & 71.94 & 77.81 & 77.56 & 77.17 \\
 & 250 & \cellcolor{gray!20}85.28 & 84.55 & 82.33 & 83.08 & 72.00 & 78.15 & 76.47 & 77.43 \\
 & 350 & \cellcolor{gray!20}85.96 & 85.05 & 83.73 & 84.12 & 72.93 & 79.24 & 78.77 & 78.17 \\
 & 450 & \cellcolor{gray!20}86.68 & 85.20 & 84.78 & 85.42 & 75.82 & 79.84 & 82.93 & 78.54 \\
\midrule
\multirow{5}{*}{ESR}
 & 100 & \cellcolor{gray!20}37.95 & 31.32 & 26.08 & 21.78 & 20.82 & 20.72 & 25.82 & 33.11 \\
 & 300 & \cellcolor{gray!20}40.30 & 36.15 & 29.53 & 22.50 & 20.70 & 17.28 & 20.95 & 37.13 \\
 & 500 & \cellcolor{gray!20}43.81 & 39.08 & 32.42 & 23.70 & 21.98 & 22.02 & 19.12 & 38.64 \\
 & 700 & \cellcolor{gray!20}44.00 & 41.27 & 35.22 & 23.58 & 22.70 & 16.98 & 19.07 & 41.75 \\
 & 900 & \cellcolor{gray!20}46.75 & 42.10 & 37.12 & 23.82 & 22.78 & 19.22 & 19.21 & 42.51 \\
\midrule
\multirow{5}{*}{GSAD}
 & 100  & \cellcolor{gray!20}91.22 & 90.50 & 48.07 & 70.34 & 21.28 & 75.80 & 40.32 & 86.22 \\
 & 300  & \cellcolor{gray!20}93.89 & 92.50 & 87.46 & 84.92 & 21.56 & 82.00 & 50.03 & 89.49 \\
 & 500  & \cellcolor{gray!20}94.08 & 93.20 & 90.77 & 90.08 & 38.10 & 88.50 & 62.23 & 90.78 \\
 & 700  & 9\cellcolor{gray!20}4.38 & 93.80 & 91.58 & 90.98 & 39.28 & 90.20 & 71.24 & 91.56 \\
 & 900  & \cellcolor{gray!20}94.66 & 94.20 & 91.92 & 89.33 & 40.21 & 91.50 & 72.01 & 92.02 \\
\midrule
\multirow{6}{*}{Letter Recognition}
 & 400  & \cellcolor{gray!20}64.33 & 60.52 & 50.28 & 59.27 & 45.80 & 55.89 & 37.25 & 56.72 \\
 & 600  & \cellcolor{gray!20}71.95 & 62.87 & 53.24 & 61.59 & 45.76 & 57.82 & 41.08 & 57.12 \\
 & 800  & \cellcolor{gray!20}73.05 & 63.78 & 56.82 & 63.27 & 46.28 & 60.13 & 42.50 & 58.82 \\
 & 1200 & \cellcolor{gray!20}76.31 & 64.95 & 59.36 & 66.13 & 46.98 & 60.75 & 44.96 & 59.13 \\
 & 1400 & \cellcolor{gray!20}77.51 & 65.21 & 60.28 & 68.20 & 49.28 & 61.03 & 46.27 & 60.23 \\
\bottomrule
\end{tabular}
\end{table*}

\begin{figure*}
    \centering
    \subfloat[sEMG]{%
        \label{fig:sub_a}%
        \includegraphics[width=0.328\textwidth]{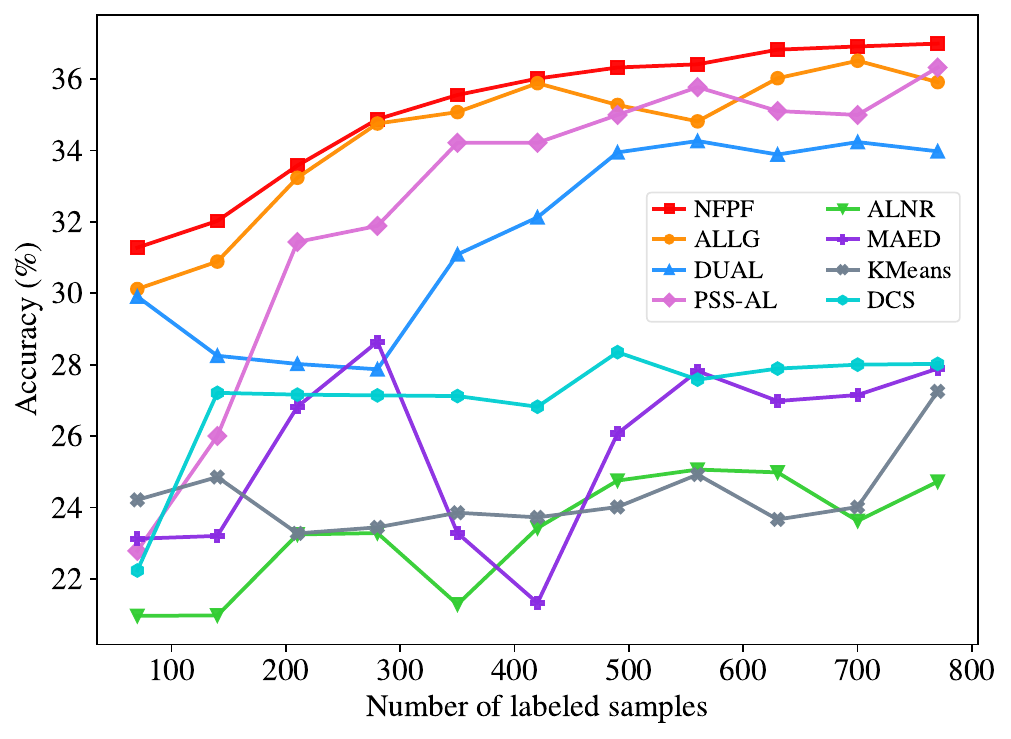}%
    }%
    \hspace{1pt} 
    \subfloat[Plant Species Leaves]{%
        \label{fig:sub_b}%
        \includegraphics[width=0.328\textwidth]{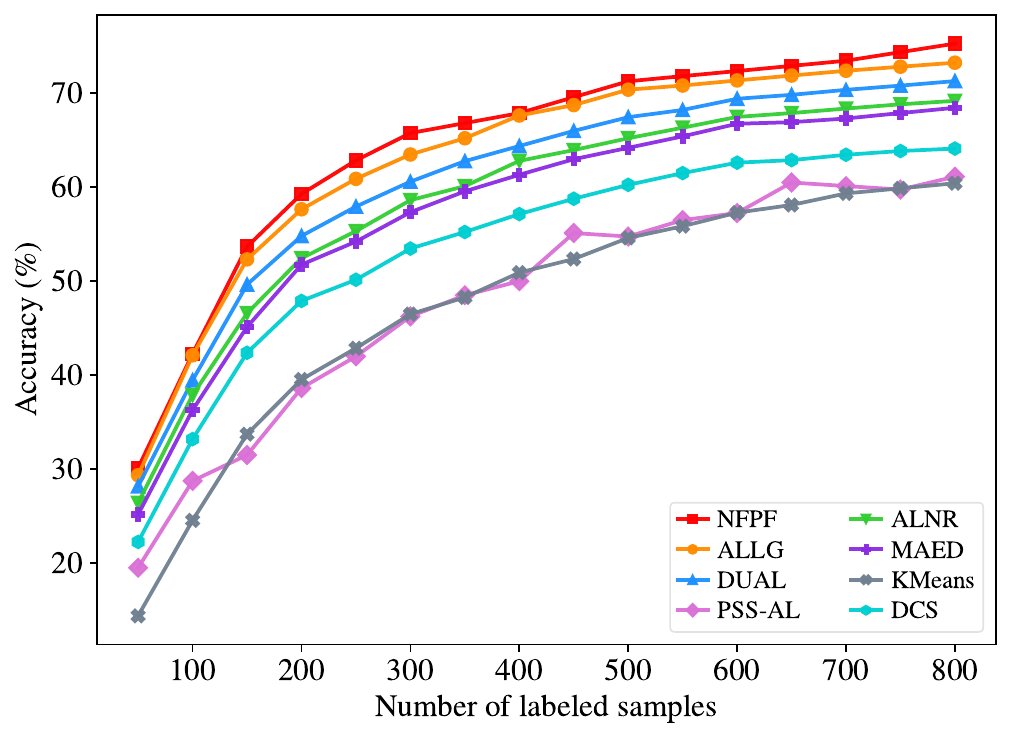}%
    }%
    \hspace{1pt}
    \subfloat[Waveform]{%
        \label{fig:sub_c}%
        \includegraphics[width=0.328\textwidth]{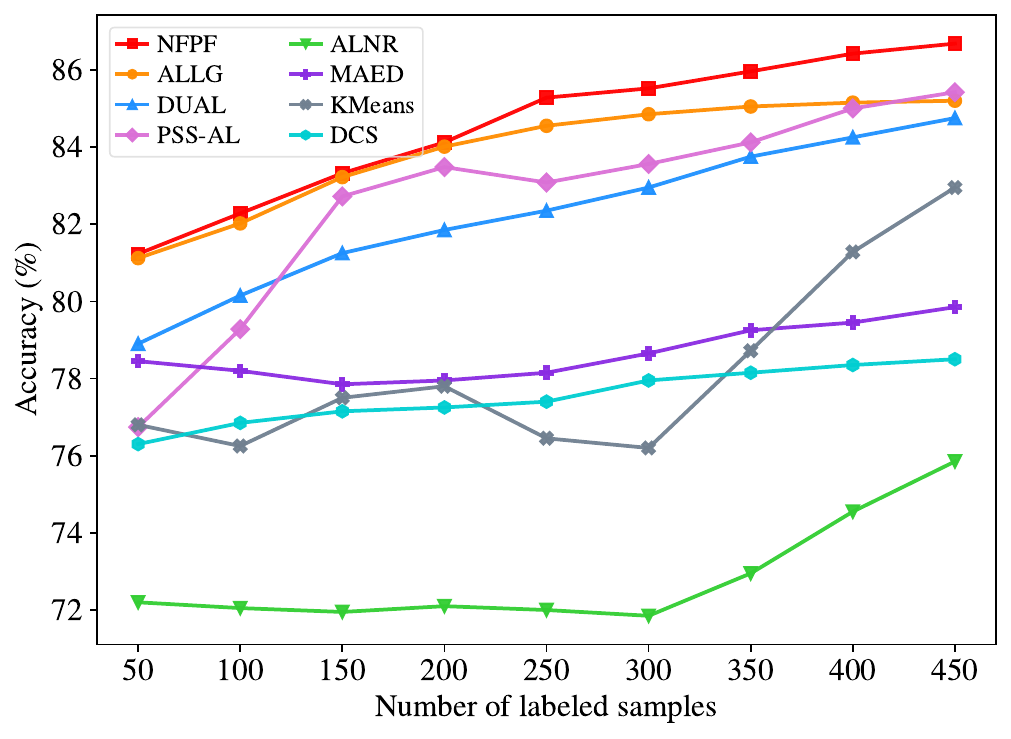}%
    }
    
    \vspace{-8pt} 
    
    \subfloat[ESR]{%
        \label{fig:sub_d}%
        \includegraphics[width=0.328\textwidth]{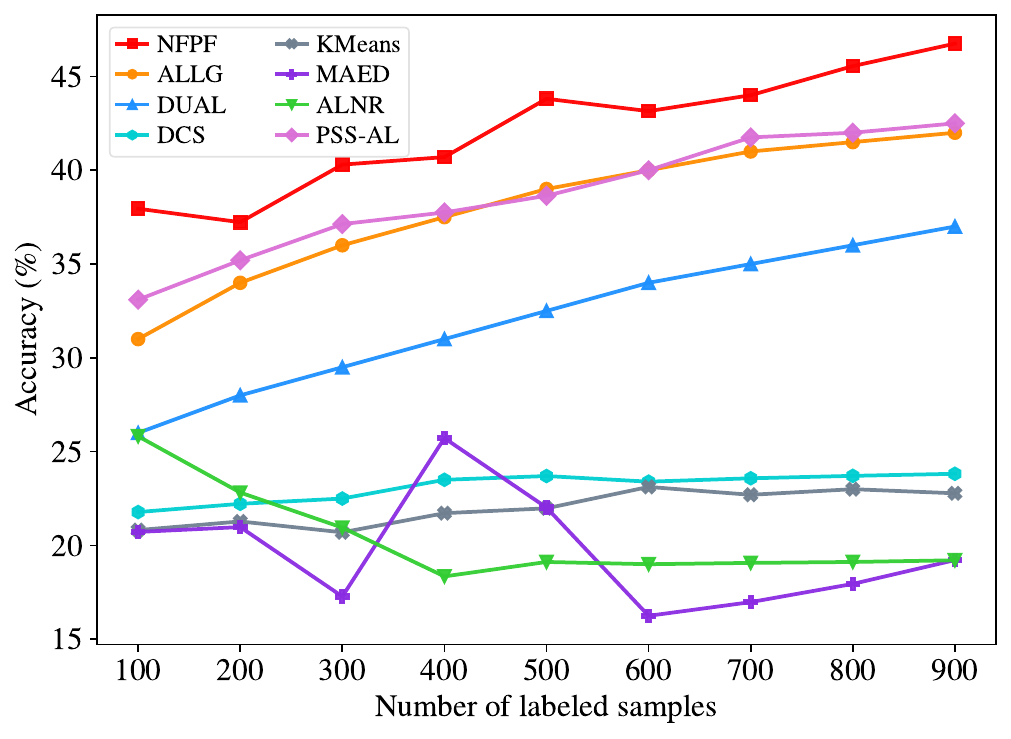}%
    }%
    \hspace{1pt}
    \subfloat[GSAD]{%
        \label{fig:sub_e}%
        \includegraphics[width=0.328\textwidth]{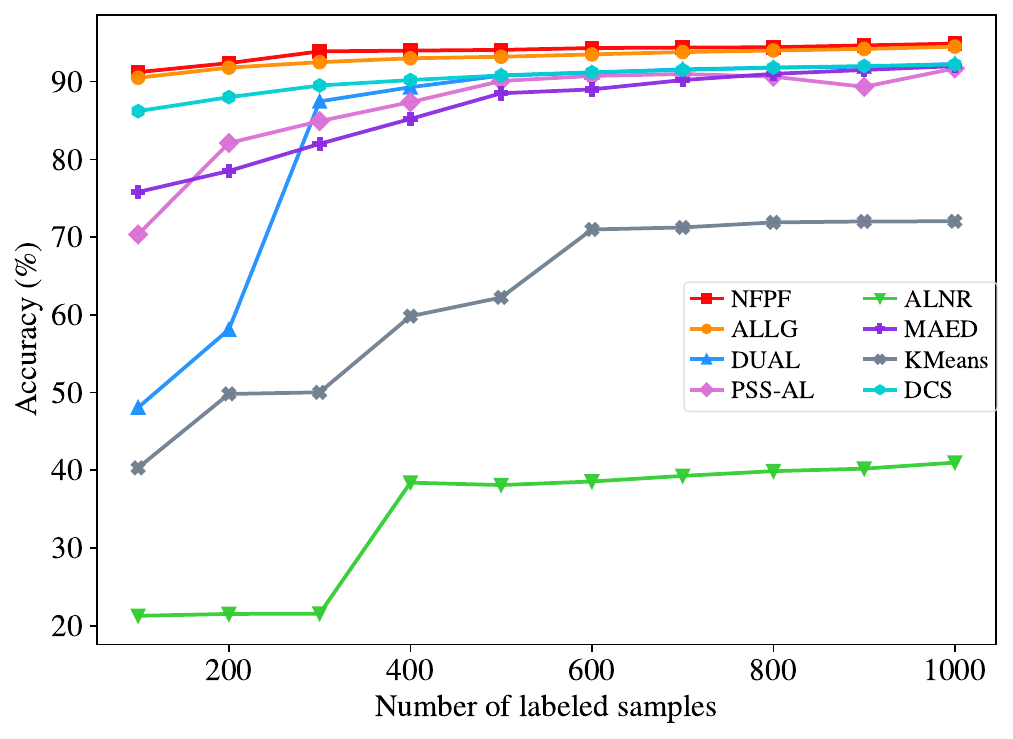}%
    }%
    \hspace{1pt}
    \subfloat[Letter Recognition]{%
        \label{fig:sub_f}%
        \includegraphics[width=0.328\textwidth]{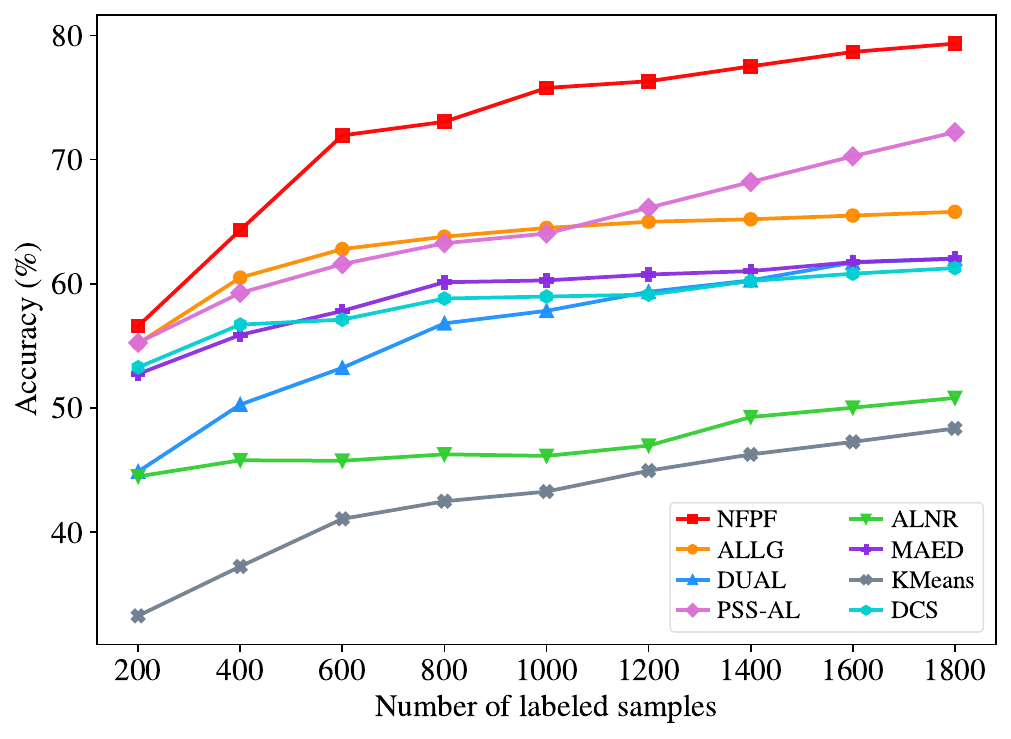}%
    }%
    \caption{Performance comparison of different methods on (a) sEMG, (b) Plant species leaves, (c) Waveform, (d) ESR, (e) GSAD, and (f) Letter recognition. Comparing methods include NFPF and other SOTAs.}
    \label{fig_UCI}
\end{figure*}

In this subsection, we focus on 6 small datasets from UCI repository. We evaluate the performance of our NFPF method and other SOTA UAL methods on six small datasets from the UCI repository, using an SVM classifier. Fig.~\ref{fig_UCI} shows the mean testing accuracy. To provide more specific metrics while conserving space, a selection of results from key datasets is presented in Table~\ref{tab:UCI}, which shows results for a representative subset of target classes. The x-axis values represent the number of labeled samples, denoted as $m$, which is the final subset size selected for the oracle to label.

The results demonstrate that NFPF consistently outperforms other methods on the ESR and Letter Recognition datasets, achieving large performance margins. The strength of NFPF on these datasets can be attributed to its ability to effectively identify both informative (boundary) and representative samples, a crucial factor for multi-class classification where a balanced selection is key. Notably, NFPF reaches a comparable performance level while requiring only one-third of the query budget m used by other approaches, highlighting its exceptional efficiency in discovering valuable data points early in the training process.

On the GSAD and Plant Species Leaves datasets, NFPF achieves performance comparable to the ALLG approach and still surpasses the other baselines. It's important to note that the overall accuracy on GSAD is inherently bounded, as the classifier reaches only about 87\% even when trained on the entire dataset. This indicates that our method's performance is approaching the theoretical maximum for this dataset, demonstrating its capability to find a high-quality subset even when the data itself poses significant classification challenges. For the sEMG dataset, NFPF demonstrates relatively stable performance as the labeling budget increases, with its average test results closely matching those of the ALLG method at query budgets of 210, 280, and 420. This stability suggests our method is resilient to the noise and complexities present in the sEMG data.

Across the six datasets, we observe that NFPF performs on par with other SOTAs when the query budget is relatively small. This is mainly because, under budget-priority settings, the size of the initial subset $\mathbf{X}_S^0$
derived from the RD needs to be carefully adjusted. This adjustment affects the overall performance of NFPF, and the impact is particularly evident in low-budget scenarios, where the method must balance selecting boundary samples while maintaining a balanced class distribution. In such cases, a small initial subset may not contain enough representative samples to properly orient the model, temporarily limiting its performance until the cumulative subset becomes more diverse.

\subsection{Experiment with Object Datasets} 
\label{OBJECT}
\begin{figure*}[htbp]
    \centering
    \subfloat[CIFAR-10]{%
        \label{fig:sub_a}%
        \includegraphics[width=0.328\textwidth]{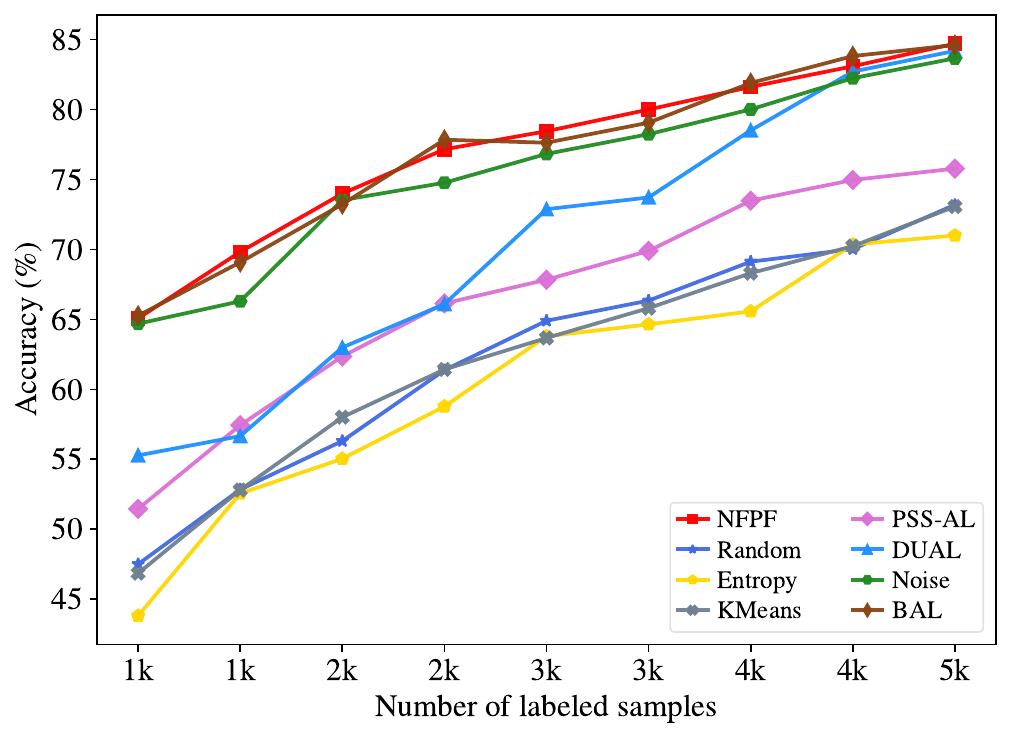}%
    }%
    \hspace{1pt} 
    \subfloat[CIFAR-100]{%
        \label{fig:sub_b}%
        \includegraphics[width=0.328\textwidth]{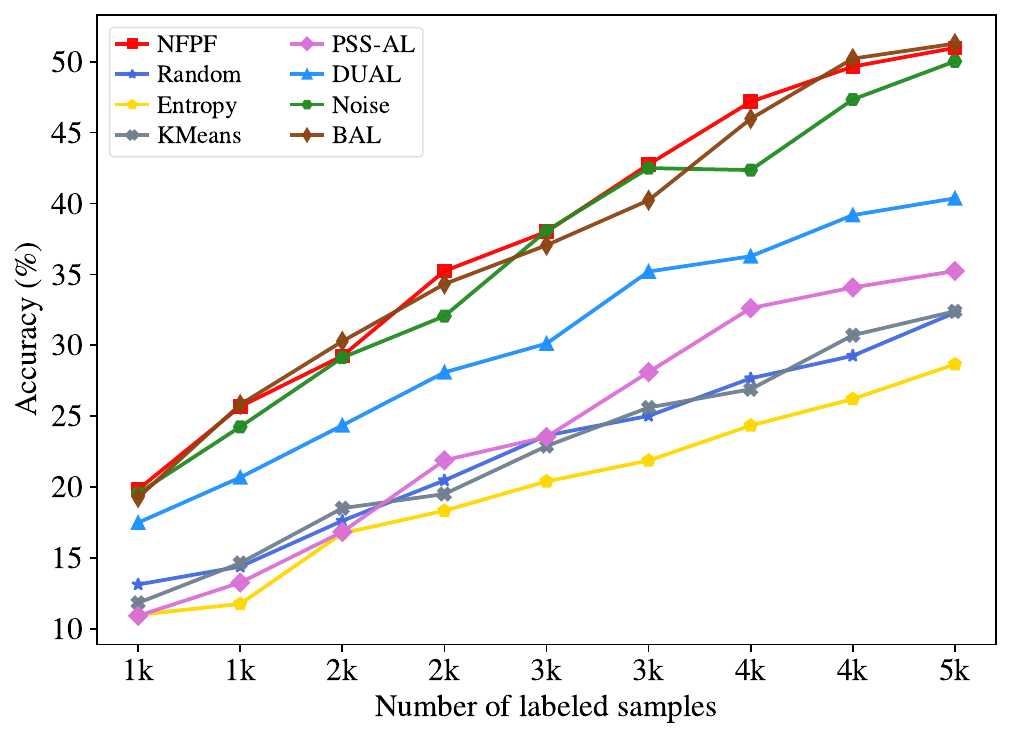}%
    }%
    \hspace{1pt} 
    \subfloat[Tiny-ImageNet]{%
        \label{fig:sub_c}%
        \includegraphics[width=0.328\textwidth]{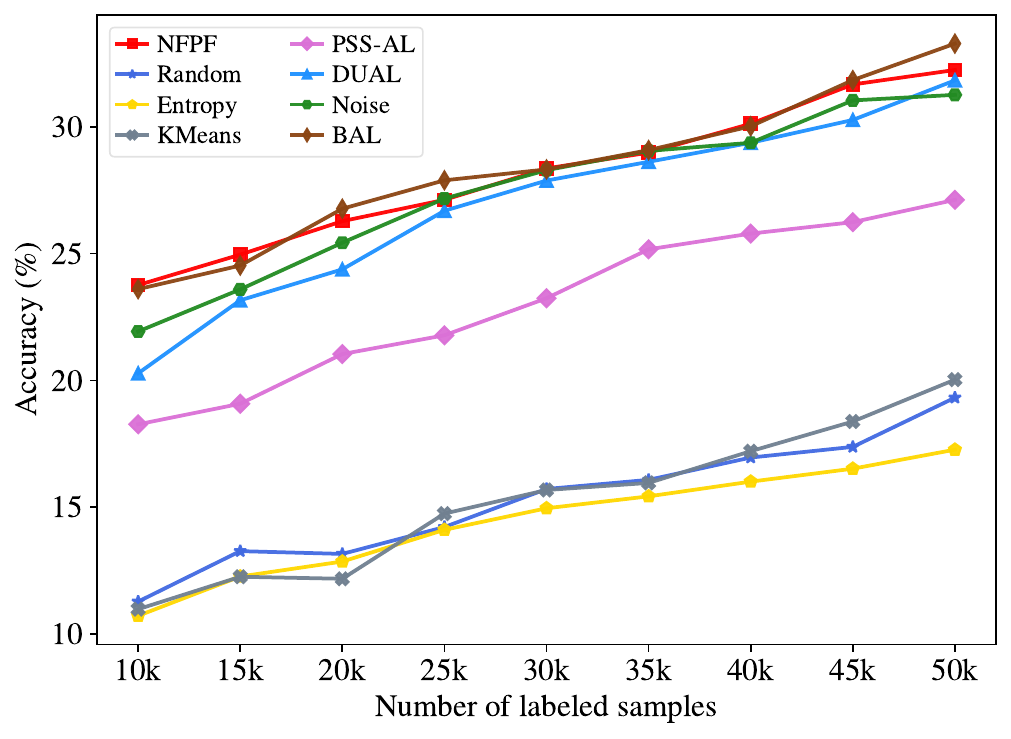}%
    }%
    \captionsetup{justification=raggedright,singlelinecheck=false}
    \caption{Performance comparison on CIFAR-10, CIFAR-100, and Tiny-ImageNet datasets.}
    \label{fig_Object}
\end{figure*}
\begin{table*}[htbp]
\centering
\begin{threeparttable}  
\caption{\centering Comparison of different methods on CIFAR-10, CIFAR-100, and Tiny-ImageNet.}
\label{tab:all_datasets_ordered}
\setlength{\tabcolsep}{6pt}
\renewcommand{\arraystretch}{1.2}
\begin{tabular}{c|c|cccccccc}
\toprule
\multirow{2}{*}{{Dataset}} & \multirow{2}{*}{Target size $m$}  & \multicolumn{8}{c}{{Method}} \\
 & & {NFPF} & {Random} & {Entropy} & {K-means} & {PSS-AL} & {DUAL} & {Noise\textsuperscript{*}} & {BAL\textsuperscript{*}} \\
\midrule
\multirow{9}{*}{CIFAR-10}
 & 1k    & \cellcolor{gray!20}65.07 & 47.51 & 43.82 & 46.86 & 51.46 & 55.28 & 64.69 & \cellcolor{cyan!10}65.26 \\
 & 1.5k  & \cellcolor{gray!20}69.82 & 52.84 & 52.56 & 52.81 & 57.44 & 56.66 & 66.29 & 69.08 \\
 & 2k    & \cellcolor{gray!20}73.98 & 56.31 & 55.04 & 58.01 & 62.35 & 62.99 & 73.52 & 73.22 \\
 & 2.5k  & \cellcolor{gray!20}77.15 & 61.37 & 58.78 & 61.42 & 66.14 & 66.09 & 74.76 & \cellcolor{cyan!10}77.83 \\
 & 3k    & \cellcolor{gray!20}78.44 & 64.90 & 63.79 & 63.67 & 67.83 & 72.87 & 76.82 & 77.62 \\
 & 3.5k  & \cellcolor{gray!20}79.99 & 66.34 & 64.64 & 65.80 & 69.89 & 73.71 & 78.23 & 79.05 \\
 & 4k    & \cellcolor{gray!20}81.61 & 69.12 & 65.57 & 68.30 & 73.48 & 78.50 & 80.00 & \cellcolor{cyan!10}81.89 \\
 & 4.5k  & \cellcolor{gray!20}83.08 & 70.02 & 70.35 & 70.23 & 74.96 & 82.70 & 82.23 & \cellcolor{cyan!10}83.81 \\
 & 5k    & \cellcolor{gray!20}84.70 & 73.18 & 70.99 & 73.06 & 75.77 & 84.18 & 83.65 & 84.60 \\
\midrule
\multirow{9}{*}{CIFAR-100}
 & 1k   & \cellcolor{gray!20}19.83 & 13.13 & 10.99 & 11.83 & 10.92 & 17.49 & 19.56 & 19.26 \\
 & 1.5k & \cellcolor{gray!20}25.68 & 14.39 & 11.76 & 14.64 & 13.26 & 20.68 & 24.25 & \cellcolor{cyan!10}25.82 \\
 & 2k   & \cellcolor{gray!20}29.25 & 17.62 & 16.74 & 18.51 & 16.84 & 24.35 & 29.13 & \cellcolor{cyan!10}30.28 \\
 & 2.5k & \cellcolor{gray!20}35.24 & 20.47 & 18.33 & 19.51 & 21.89 & 28.09 & 32.06 & 34.32 \\
 & 3k   & \cellcolor{gray!20}38.01 & 23.64 & 20.40 & 22.89 & 23.54 & 30.12 & 38.08 & 37.06 \\
 & 3.5k & \cellcolor{gray!20}42.76 & 25.02 & 21.87 & 25.61 & 28.11 & 35.20 & 42.50 & 40.23 \\
 & 4k   & \cellcolor{gray!20}47.18 & 27.67 & 24.34 & 26.90 & 32.62 & 36.28 & 42.36 & 45.98 \\
 & 4.5k & \cellcolor{gray!20}49.66 & 29.26 & 26.21 & 30.71 & 34.08 & 39.18 & 47.34 & \cellcolor{cyan!10}50.22 \\
 & 5k   & \cellcolor{gray!20}50.98 & 32.30 & 28.66 & 32.40 & 35.24 & 40.34 & 50.03 & \cellcolor{cyan!10}51.28 \\
\midrule
\multirow{9}{*}{Tiny-ImageNet}
 & 10k  & \cellcolor{gray!20}23.76 & 11.28 & 10.72 & 10.98 & 18.27 & 20.28 & 21.93 & 23.60 \\
 & 15k  & \cellcolor{gray!20}24.96 & 13.27 & 12.27 & 12.26 & 19.08 & 23.16 & 23.59 & 24.53 \\
 & 20k  & \cellcolor{gray!20}26.28 & 13.16 & 12.86 & 12.18 & 21.04 & 24.38 & 25.43 & \cellcolor{cyan!10}26.77 \\
 & 25k  & \cellcolor{gray!20}27.12 & 14.22 & 14.11 & 14.75 & 21.78 & 26.69 & 27.17 & \cellcolor{cyan!10}27.89 \\
 & 30k  & \cellcolor{gray!20}28.36 & 15.72 & 14.96 & 15.68 & 23.24 & 27.88 & 28.29 & 28.32 \\
 & 35k  & \cellcolor{gray!20}28.98 & 16.08 & 15.43 & 15.96 & 25.17 & 28.62 & 29.05 & \cellcolor{cyan!10}29.08 \\
 & 40k  & \cellcolor{gray!20}30.12 & 16.96 & 16.01 & 17.21 & 25.79 & 29.37 & 29.37 & 30.02 \\
 & 45k  & \cellcolor{gray!20}31.67 & 17.38 & 16.52 & 18.38 & 26.24 & 30.27 & 31.04 & \cellcolor{cyan!10}31.84 \\
 & 50k  & \cellcolor{gray!20}32.24 & 19.32 & 17.27 & 20.03 & 27.12 & 31.83 & 31.26 & \cellcolor{cyan!10}33.28 \\
\bottomrule
\end{tabular}
\begin{tablenotes}
    \footnotesize
    \item [*]: Represents supervised active learning method.
\end{tablenotes}
\end{threeparttable}
\end{table*}

In this subsection, to further validate the effectiveness of our method, we conduct experiments on CIFAR-10, CIFAR-100, and Tiny-ImageNet. Results are shown in Fig.~\ref{fig_Object} and illustrate in Table ~\ref{tab:all_datasets_ordered}. The grey block in Table~\ref{tab:all_datasets_ordered} highlights the best results achieved between the supervised active learning method Noise and the other UAL approaches, whereas the blue block denotes cases where the supervised active learning method BAL outperforms our approach. We compare the performance of various techniques under different target proportions $m$. The results demonstrate that Noise and BAL achieve superior performance relative to other methods, while our NFPF attains performance on par with the latter.

As the size of the target subset increases, the performance of our NFPF approach improves correspondingly. Our algorithm demonstrates a faster rate of performance improvement and a higher overall performance compared to UAL and other baseline algorithms. This indicates that the subsets selected by NFPF enable downstream classification tasks to converge more quickly and achieve higher accuracy. Rather than relying on exhaustive annotation, NFPF identifies a subset of data whose informativeness rivals that of a fully labeled set under comparable budgets. By training exclusively on this carefully curated subset, nearly 25\% of annotation effort can be spared while preserving performance integrity.

Furthermore, our NFPF method achieves performance that is comparable to supervised Active Learning methods like Bal and Noise. While a performance gap still exists between NFPF and Bal, as highlighted in the blue-colored sections of Table~\ref{tab:all_datasets_ordered}, our approach offers significant practical advantages. As an unsupervised algorithm, it eliminates the need for extensive manual annotation during iterative training, making it far more effective in real-world applications where labeling resources are scarce.

\subsection{Visualization} 
\label{visual}
\begin{figure*}
    \centering
    \subfloat[K-Means]{%
        \includegraphics[width=0.16\textwidth]{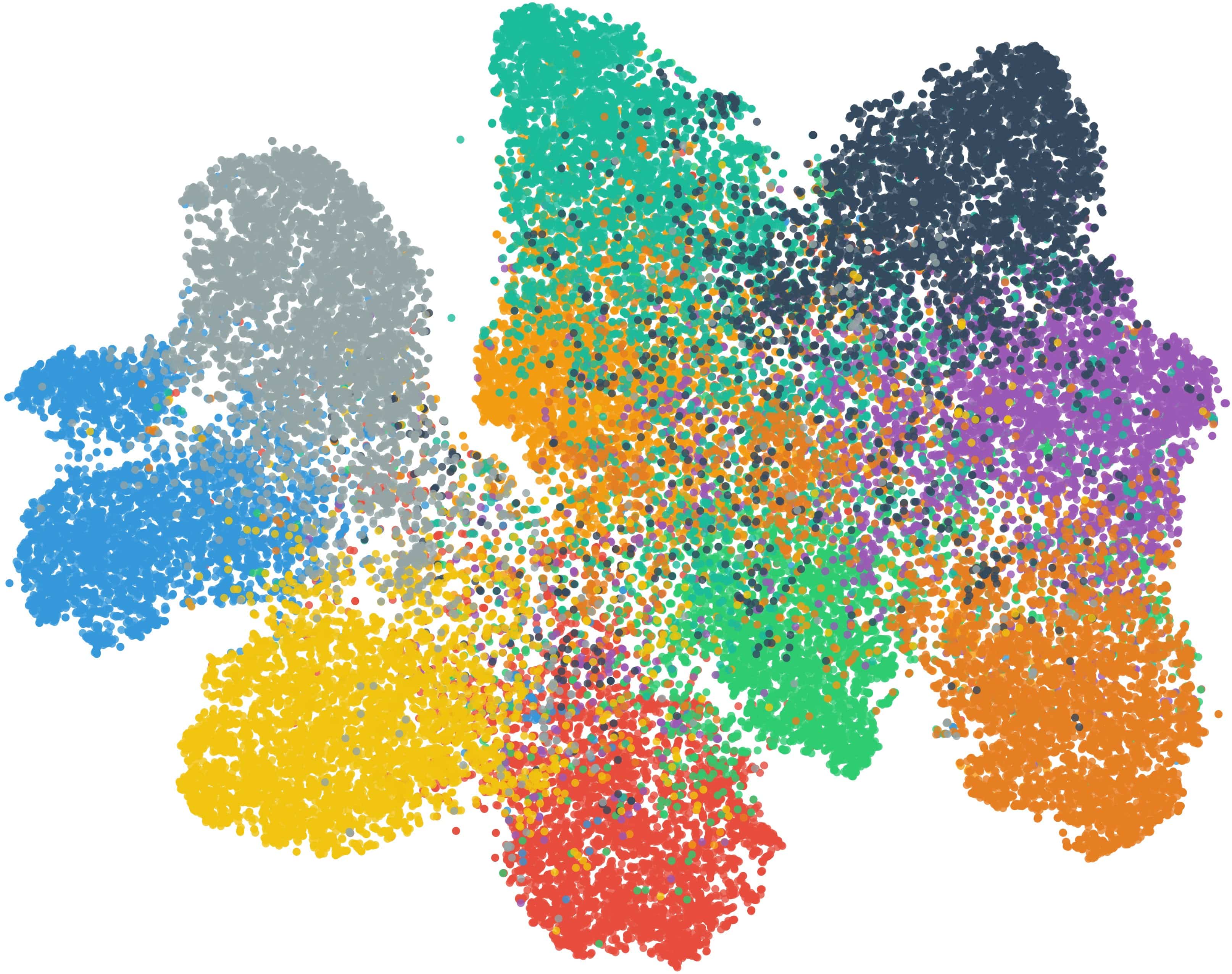}%
    }\hspace{1pt}
    \subfloat[Entropy]{%
        \includegraphics[width=0.16\textwidth]{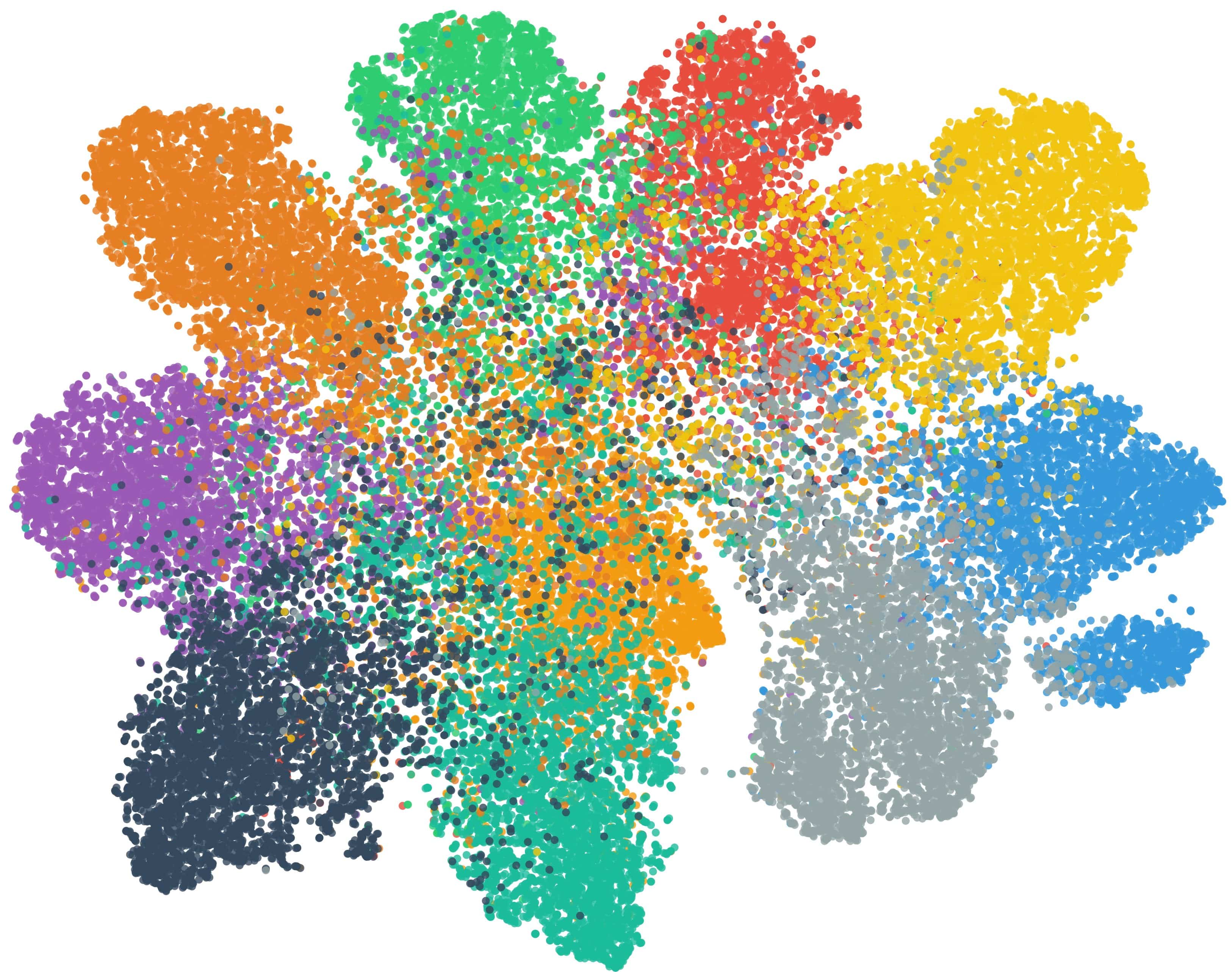}%
    }\hspace{1pt}
    \subfloat[Random]{%
        \includegraphics[width=0.16\textwidth]{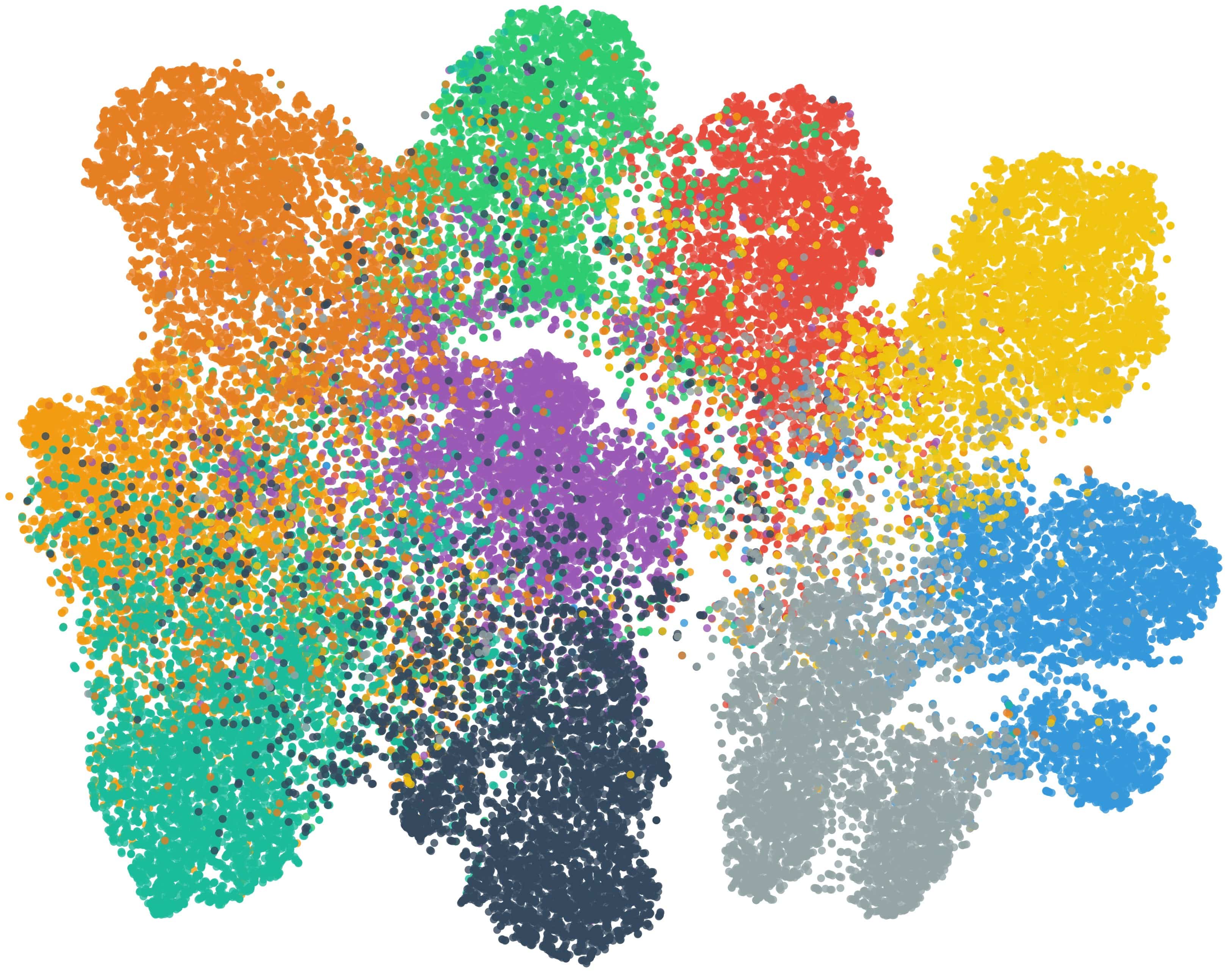}%
    }\hspace{1pt}
    \subfloat[PSS-AL]{%
        \includegraphics[width=0.16\textwidth]{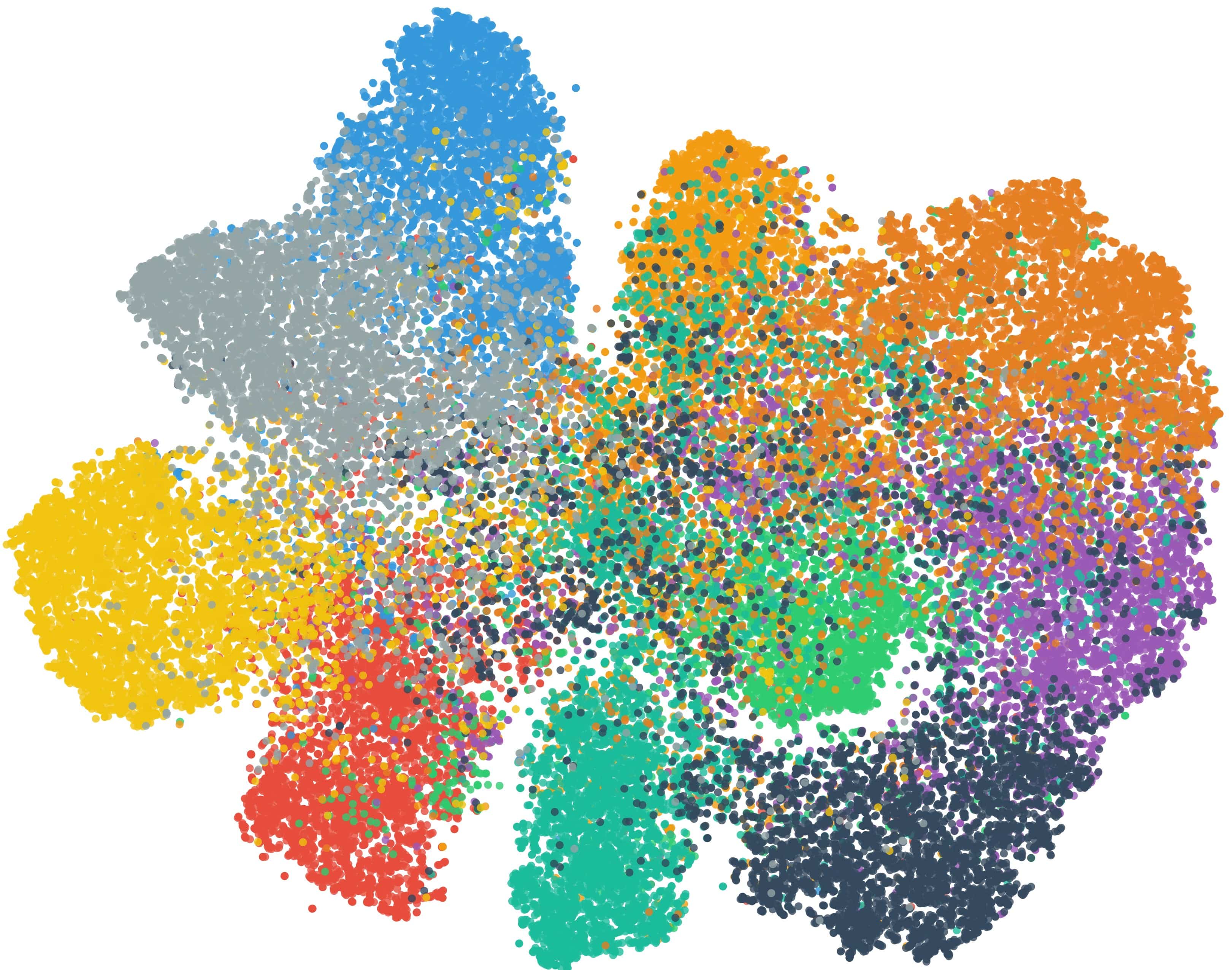}%
    }\hspace{1pt}
    \subfloat[DUAL]{%
        \includegraphics[width=0.16\textwidth]{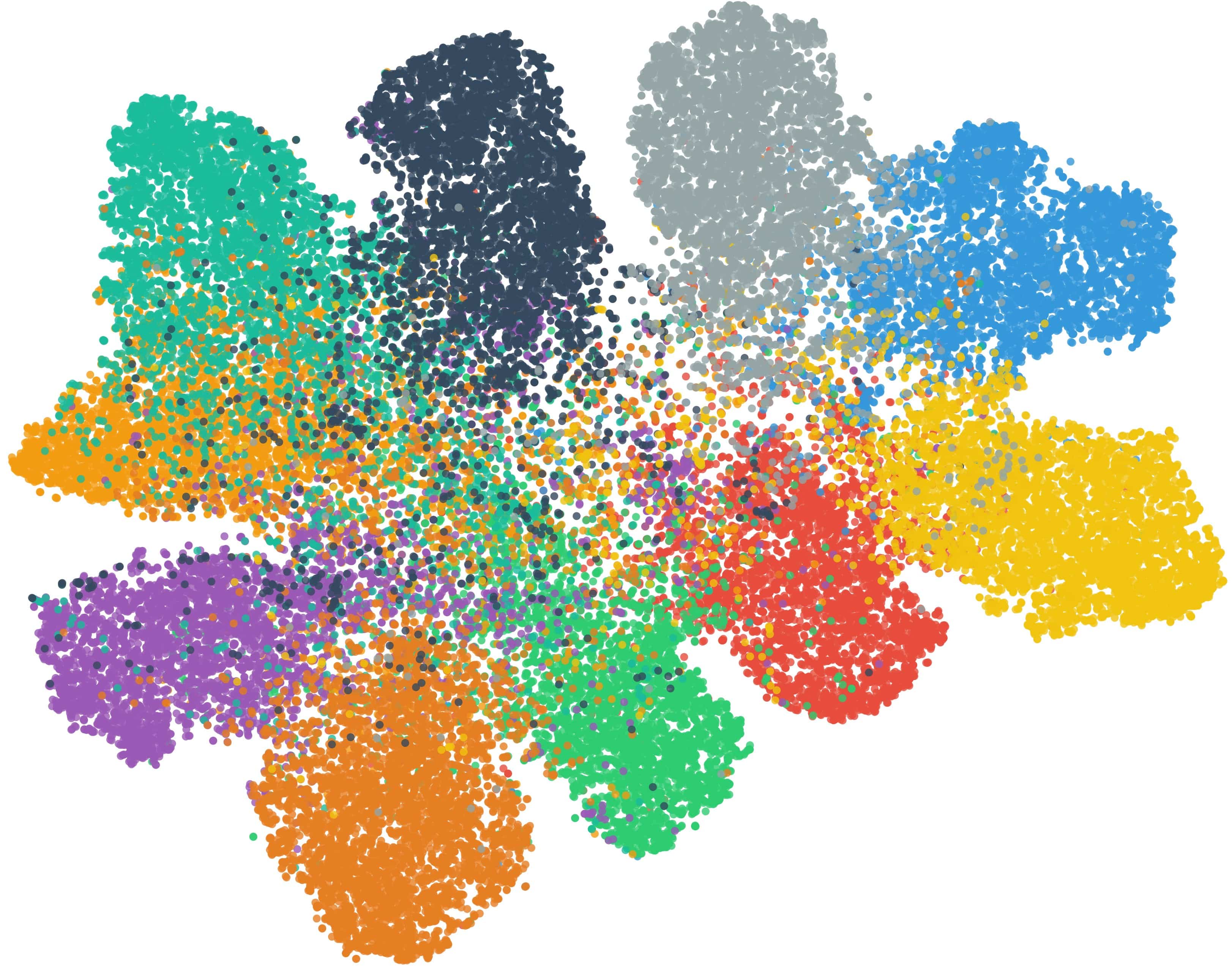}%
    }\hspace{1pt}
    \subfloat[NFPF]{%
        \includegraphics[width=0.15\textwidth]{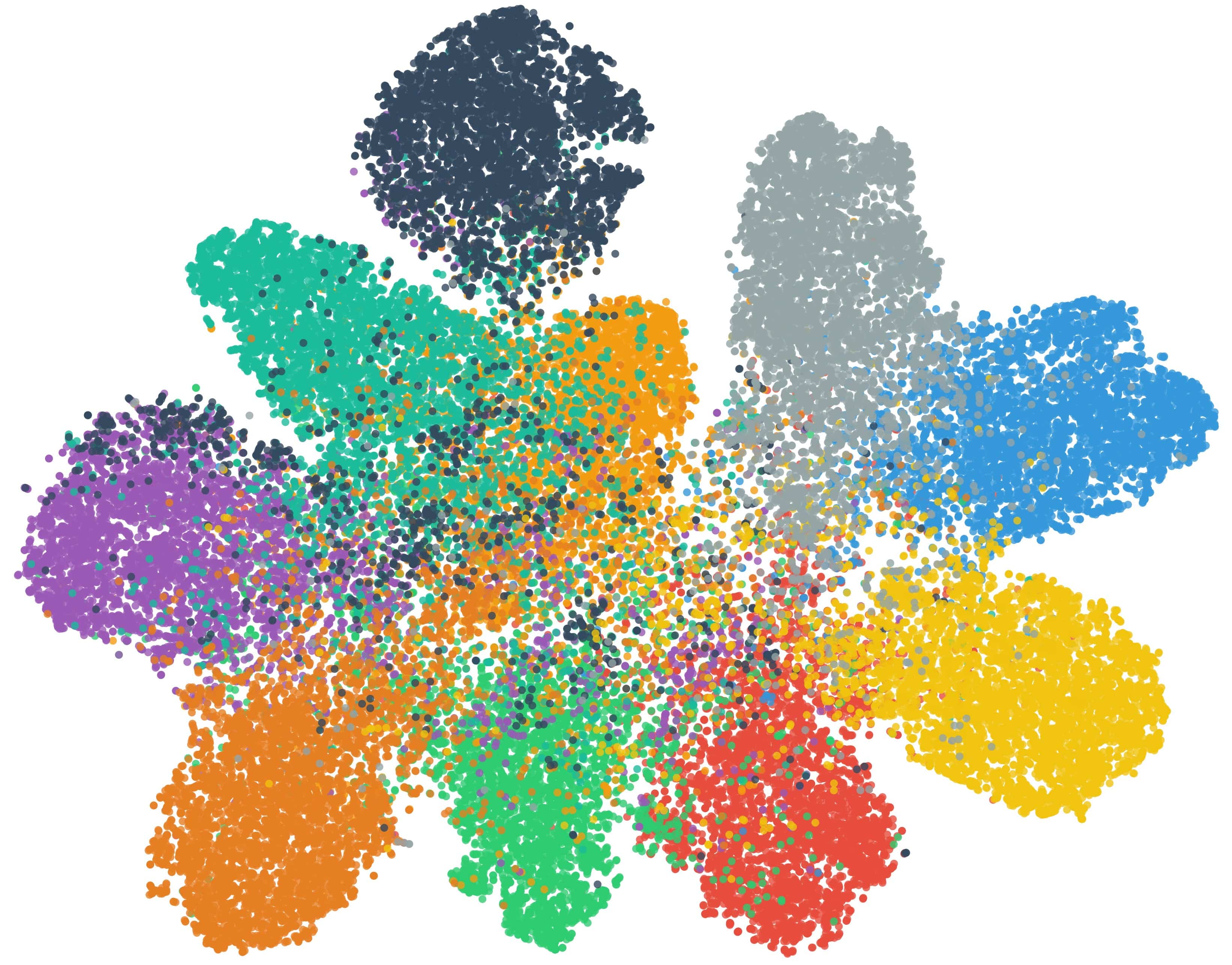}%
    }
    \caption{ t-SNE visualization of features learned by (a) NFPF, (b) K-Means, (c) Entropy, (d) Random, (e) PSS-AL, (f) DUAL on CIFAR-10. Different colors represent labeled samples.}
    \label{fig_t_SNE}
\end{figure*}

\begin{figure}   
  \centering
  \includegraphics[width=0.95\linewidth]{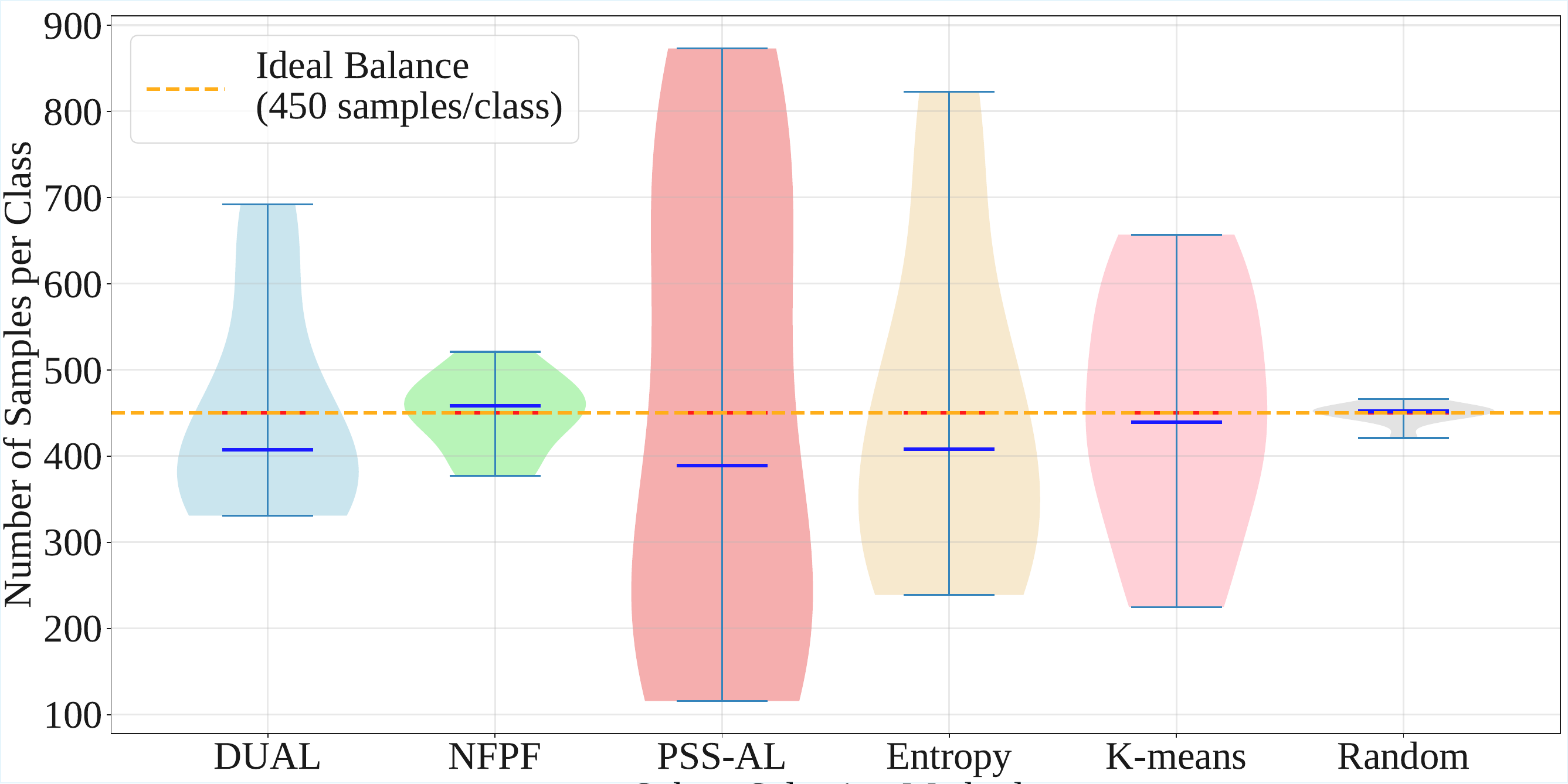}
  \caption{Comparison of class distributions in the subsets selected by different methods on CIFAR-10.
}
  \label{fig:distribution}
\end{figure}

In this subsection, we set the target subset size to $\mathbf{X}_S=4500$ on CIFAR-10, and train ResNet-18 with 70 learning epochs using subsets obtained from different approaches. he resulting model weights are then used to extract features, which are visualized via t-SNE for a more intuitive illustration, as shown in Fig.~\ref{fig_t_SNE}. The visualization clearly reveals the quality of the subsets chosen by each method. When using baseline approaches like K-Means, Entropy, or Random sampling, features from different categories tend to show significant overlap. This overlap indicates that the selected subsets are not sufficiently informative to guide the model toward learning well-separated classes, resulting in a cluttered and less discriminative feature space.
In contrast, methods like PSS-AL and DUAL show noticeable improvements over these simple baselines, as their feature embeddings display a better degree of class separation. However, our NFPF method stands out. The visualization of our NFPF-selected subset reveals a feature space where classes are separated with notably clearer and more distinct boundaries. This is particularly evident on the right side of the t-SNE plot, where our method forms tightly clustered, non-overlapping groups. This superior class separation provides compelling visual evidence that our NFPF approach effectively identifies and selects the most valuable samples for training, thereby enabling the model to learn a more robust and discriminative feature representation.

Fig.~\ref{fig:distribution} illustrates the category distributions of the samples selected by different approaches. Our NFPF method achieves a more balanced class distribution in its selections compared to other methods, a property that is only matched by random sampling. However, a key distinction must be made: while random sampling is inherently class-agnostic, its lack of an informative selection criterion compromises its ability to yield optimal model performance. Prior studies, such as \cite{yi2022pt4al}, have demonstrated that distributional imbalance can adversely affect performance. This is particularly relevant in active learning, where both informativeness and representativeness are crucial. As we discussed, while selecting boundary samples provides strong informativeness, maintaining balanced class coverage is equally critical. Our method's ability to achieve a harmonious balance between these two properties—identifying highly informative samples while maintaining a balanced distribution—is a key factor contributing to its superior and more robust performance.

\section{ABLATION EXPERIMENTS} 
\begin{figure*}[tbp]
    \centering
    \subfloat[SVM on Waveform]{%
        \label{fig:W}%
        \includegraphics[width=0.32\textwidth]{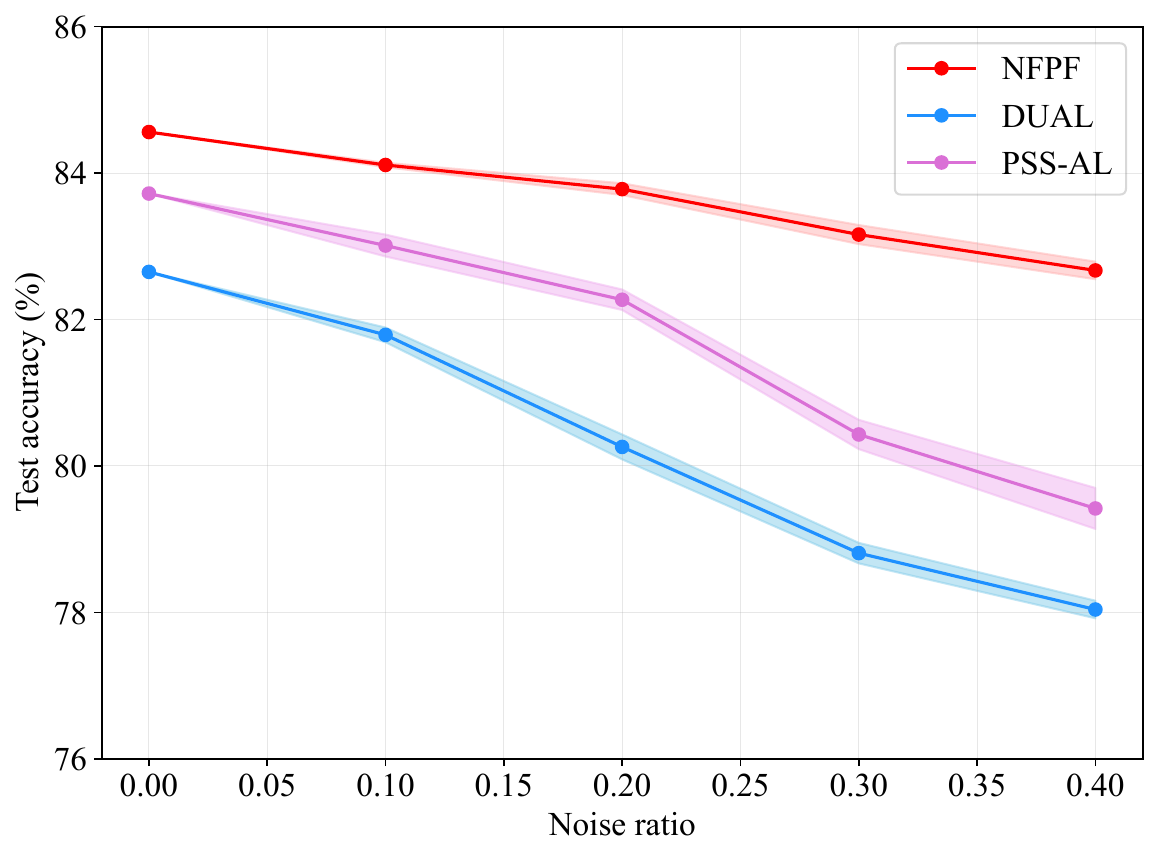}%
    }%
    \hspace{1pt} 
    \subfloat[ResNet-18 on CIFAR-10]{%
        \label{fig:10}%
        \includegraphics[width=0.32\textwidth]{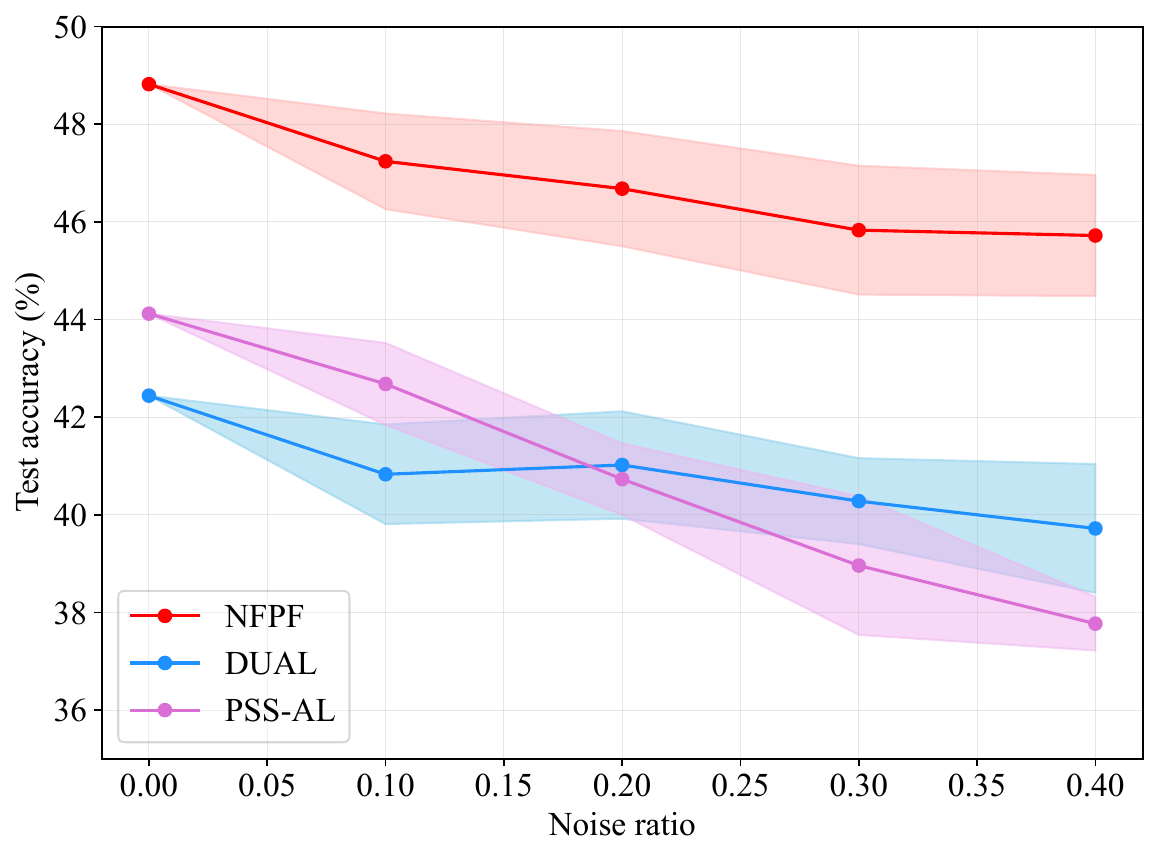}%
    }%
    \hspace{1pt} 
    \subfloat[ResNet-18 on CIFAR-100]{%
        \label{fig:100}%
        \includegraphics[width=0.32\textwidth]{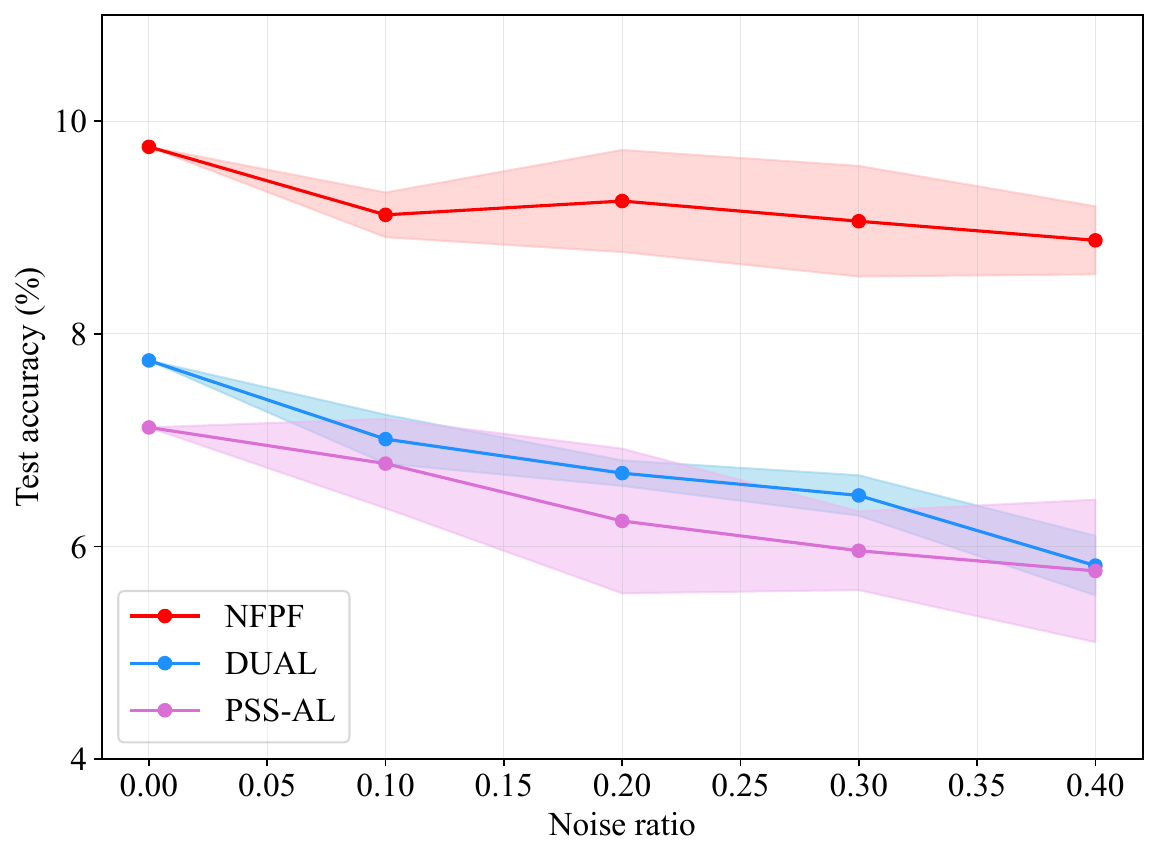}%
    }%
    \captionsetup{justification=raggedright,singlelinecheck=false}
    \caption{Test accuracy under different noise ratios. The shaded area represents the error range.}
    \label{fig_noise}
\end{figure*} 
\label{Ablation}
The purpose of this is to study the sensitivity of involved main parameter and the robustness under data noise in our proposed framework by carrying on multiple internal experiments. Fair tests will be applied on three datasets, i.e, Waveform, CIFAR-10 and CIFAR-100.

\subsection{Performance Under Label Noise} 
We assess our method's performance under label noise to evaluate its resilience. Specifically, we measure its data summarization capabilities on the Waveform, CIFAR-10, and CIFAR-100 datasets. The experimental setup and evaluation methodology remain consistent with those detailed in Section~\ref{UCI} and Section ~\ref{OBJECT}. 
We evaluate our method's performance under noisy conditions by applying symmetric label noise to randomly corrupt samples at different ratios. This corrupted dataset, which we denote as $\mathbf{X}$, is then used to select a target subset $\mathbf{X}_S$ containing 220 samples. The comparison is conducted against DUAL and PSS-AL. It is important to note that the initial subset $\mathbf{X}_S^0$ is also chosen directly from the noisy dataset.

\begin{figure*}
    \centering
    \subfloat[SVM on Waveform]{%
        \label{fig:W}%
        \includegraphics[width=0.32\textwidth]{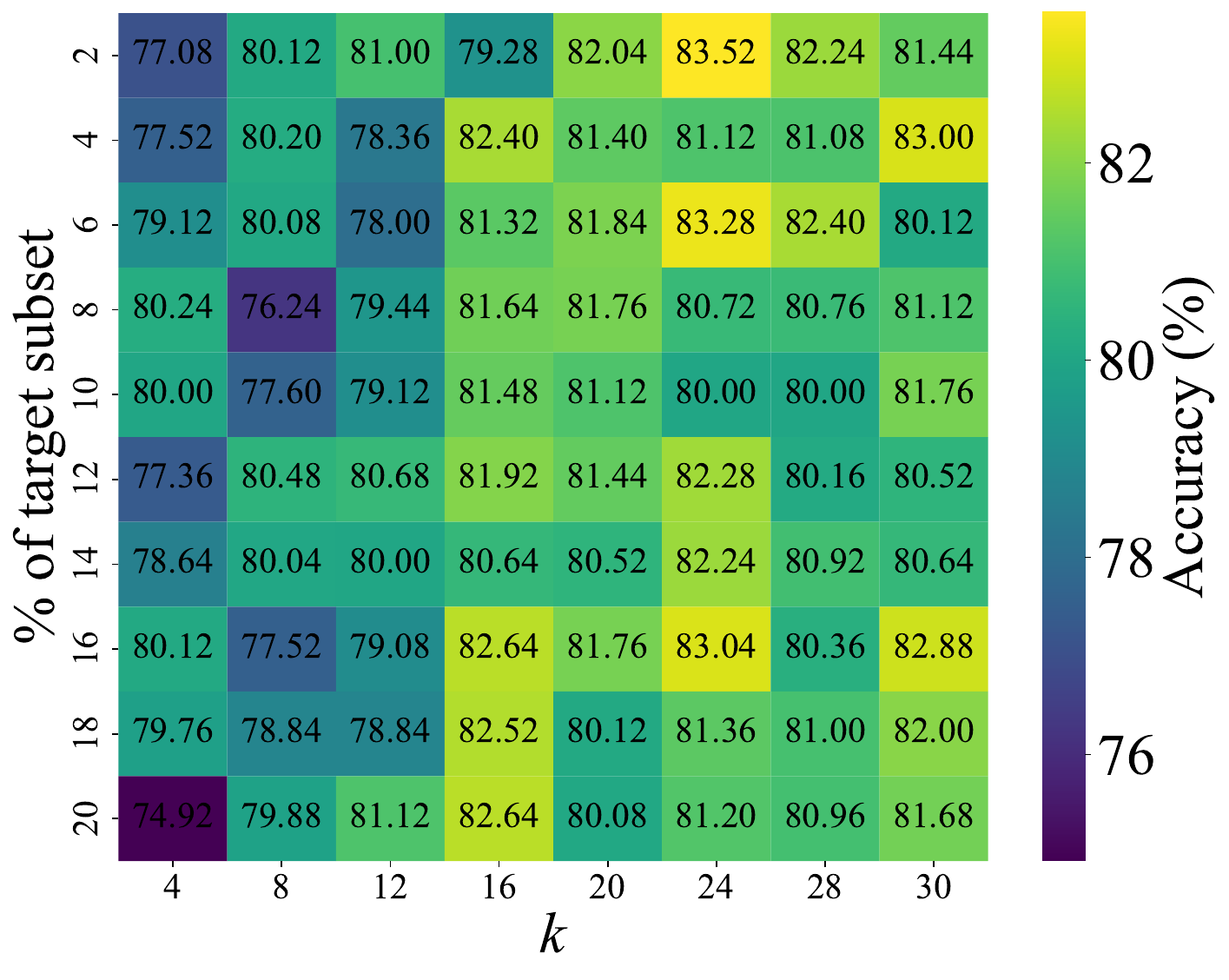}%
    }%
    \hspace{1pt} 
    \subfloat[ResNet-18 on CIFAR-10]{%
        \label{fig:10}%
        \includegraphics[width=0.32\textwidth]{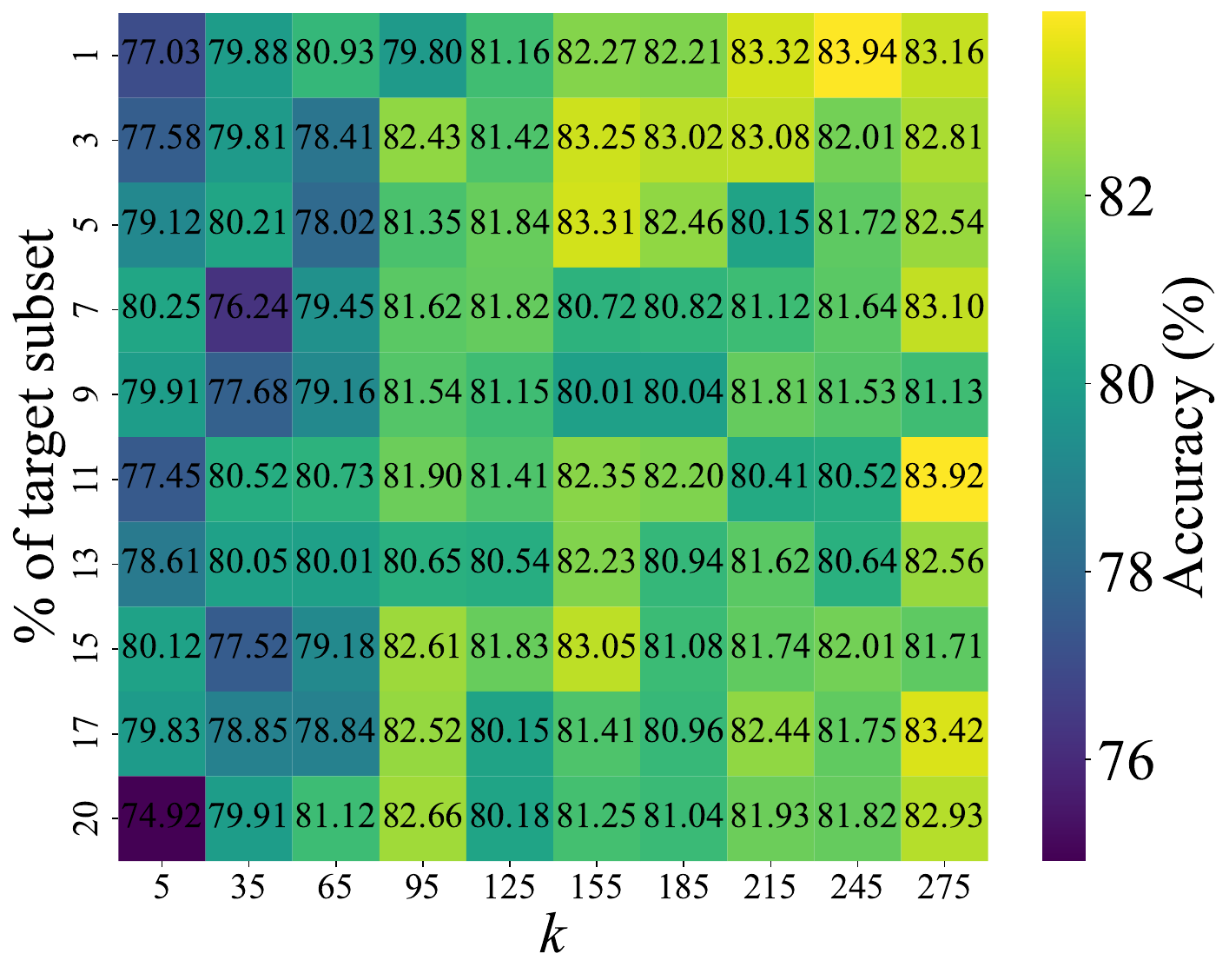}%
    }%
    \hspace{1pt} 
    \subfloat[ResNet-18 on CIFAR-100]{%
        \label{fig:100}%
        \includegraphics[width=0.32\textwidth]{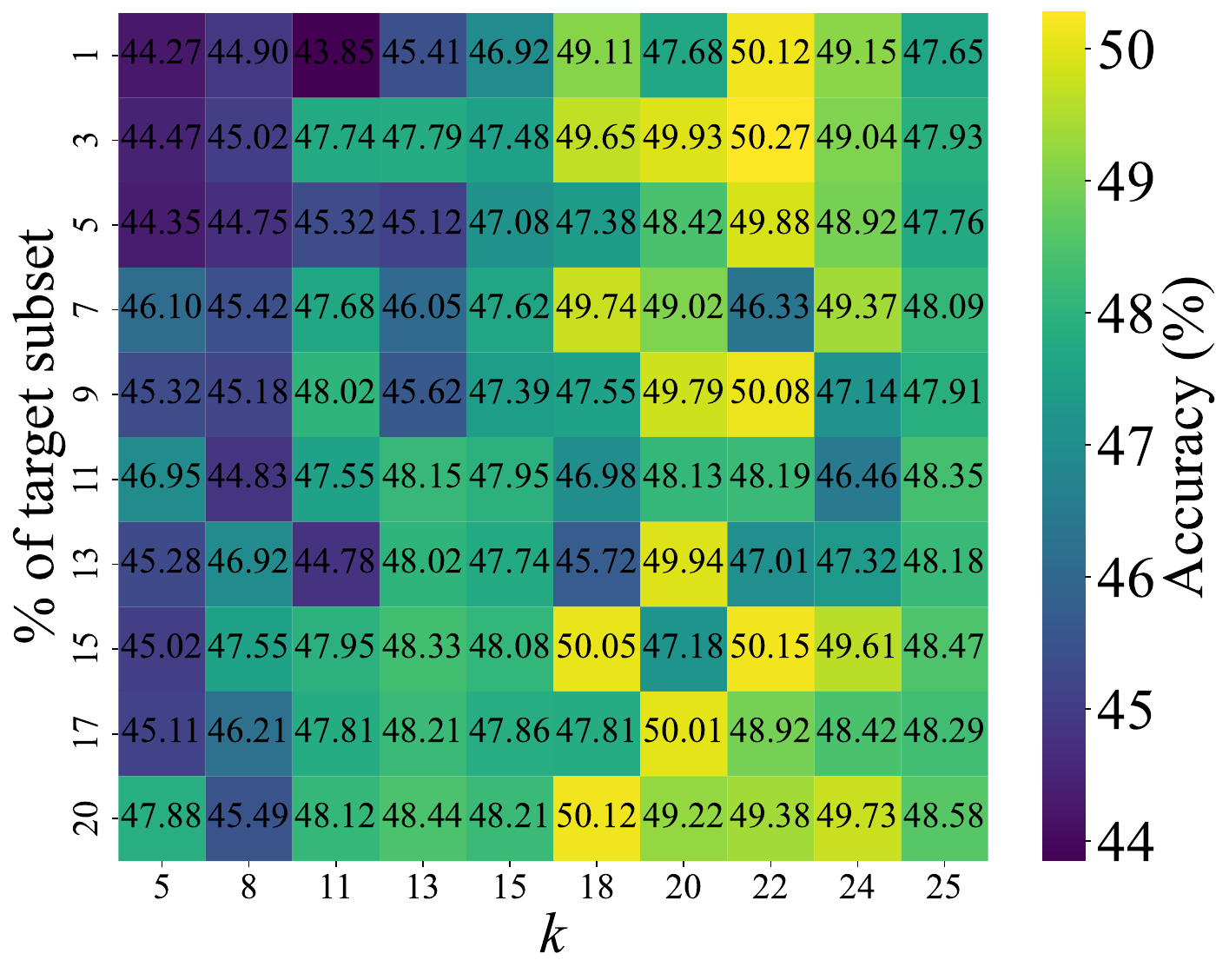}%
    }%

    \caption{ Test accuracy with different parameters settings. The x-axis indicates the initial subset size $k$, and the y-axis represents the number of selected samples per cycle $n$, normalized by the target subset size $m$.}
    \label{fig_paraMeter}
\end{figure*} 
As shown in Fig.~\ref{fig_noise}, our method consistently outperforms other baselines in noisy conditions. 

For the Waveform task, NFPF's performance remains relatively stable even as the noise ratio increases. In a no-noise setting, NFPF achieves an accuracy of approximately 84.5\%, while DUAL and PSS-AL are at about 82\% and 79\%, respectively. At a high noise ratio of 0.40, NFPF's accuracy still holds at around 81\%, whereas DUAL and PSS-AL drop to about 78\% and 76.5\%, respectively. NFPF consistently maintains a significant performance margin.

For the CIFAR-10 task, our NFPF method demonstrates exceptional resilience to label noise. As the noise ratio increases, NFPF's accuracy experiences a relatively gradual decline, dropping from 48.4\% to 46.0\%. While the DUAL method also shows comparable robustness with a slow performance drop from 42.2\% to 39.9\%, our NFPF consistently maintains a notable performance advantage across all noise levels.

For the CIFAR-100 task, the NFPF method exhibits the most gradual performance decline, indicating a high degree of stability. Furthermore, its performance fluctuations across different noise levels are relatively smaller than those of the PSS-AL method, as evidenced by its lower variance in test results.

\subsection{Sensitivity of Parameters} 
In this section, we adopt the same parameter settings as described in Section~\ref{visual} for CIFAR-10 and CIFAR-100. Additionally, we introduce Wavform on a scale of $\mathbf{X}_S=150$ to further validate the impact of our parameters.

To analyze the impact of the initial subset size, $k$, and the number of selected samples in each cycle, $n$, we employed a mixed validation strategy. The parameter $k$ is crucial as it dictates the training of the initial SFLM model in the first cycle. A thoughtfully chosen initial subset, $\mathbf{X}_S^0$, can provide the current SFLM model with a robust representation of local features by including samples located near the decision boundary.

Meanwhile, the number of selected samples per cycle, $n$, determines the number of cycles required to reach the target subset size, $m$. A larger $n$ selects more samples in each iteration, which can, to some extent, improve training efficiency. However, this may introduce noisy samples and prevent the current model from learning more discriminative features effectively. In contrast, a smaller $n$ increases the number of iterations, leading to longer training times but often yielding better performance. The heatmap illustrating the performance of different parameter combinations is presented in Fig.~\ref{fig_paraMeter}. 

\begin{figure}   
  \centering
  \includegraphics[width=0.95\linewidth]{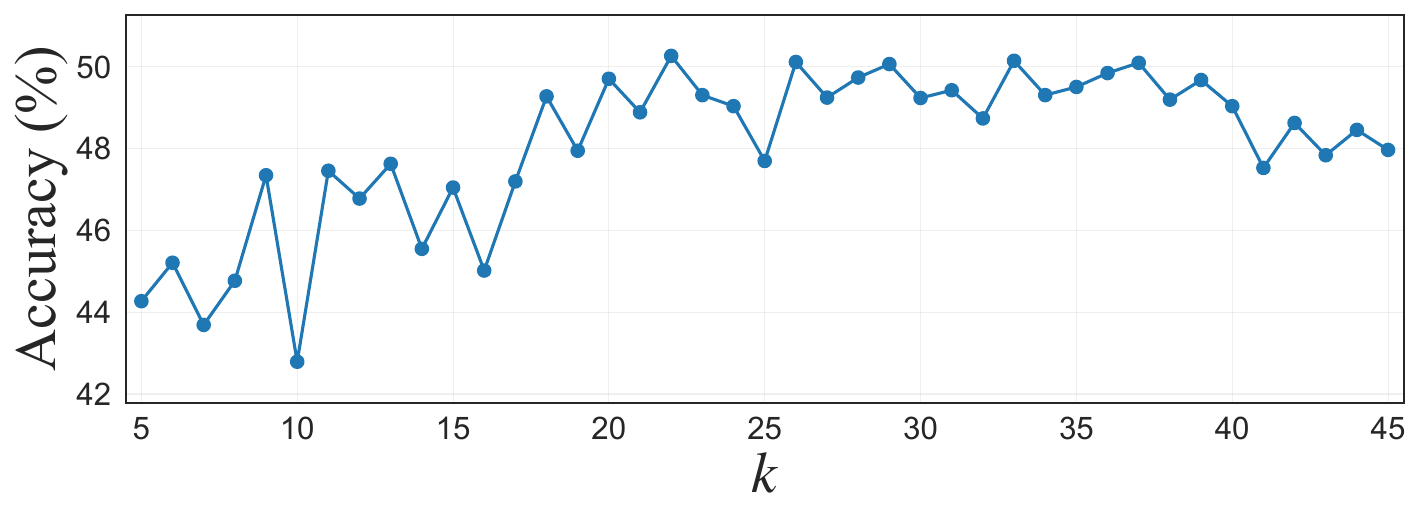}
  \caption{Impact of parameter $k$ on CIFAR-100 classification accuracy.
}
  \label{fig:para_cifar100}
\end{figure}

The parameter analysis reveals that the influence of parameters $n$ and $k$ varies across different datasets. We found that the optimal scale for $k$ typically falls within the range of $30-50\%$. This demonstrates the effectiveness of our proposed RD algorithm, which successfully selects informative samples to a certain extent. However, performance degrades as RD increases. We validated this observation with a separate set of experiments on CIFAR-100 in Fig.\ref{fig:para_cifar100}, where we independently varied the value of $k$. In contrast, the setting of parameter $n$ appears to have no significant impact on overall performance. Experimental analysis shows that a smaller value for $n$ generally leads to more stable and better overall performance. 

\section{CONCLUSION} 
\label{conclu}
In this work, we proposed a novel unsupervised active learning framework, the NFPF, to ad dress the core challenges of current UAL methods. By introducing the SFLM and the RD metric, we have proposed a new paradigm for measuring sample informativeness that moves beyond the limitations of gradient-based scoring. Our experimental results prove that NFPF achieves remarkable performance, outclassing existing UAL methods and closing the gap with supervised Active Learning on diverse vision datasets. The findings from our ablation studies and visualizations confirm that NFPF's success is due to its robust ability to select a representative and informative subset of data. This work paves the way for future research into unsupervised sample selection, with potential applications in domains where labeling is prohibitively expensive.

\bibliographystyle{IEEEtran}
\bibliography{references}

\begin{thebibliography}{10}
\providecommand{\url}[1]{#1}
\csname url@samestyle\endcsname
\providecommand{\newblock}{\relax}
\providecommand{\bibinfo}[2]{#2}
\providecommand{\BIBentrySTDinterwordspacing}{\spaceskip=0pt\relax}
\providecommand{\BIBentryALTinterwordstretchfactor}{4}
\providecommand{\BIBentryALTinterwordspacing}{\spaceskip=\fontdimen2\font plus
\BIBentryALTinterwordstretchfactor\fontdimen3\font minus \fontdimen4\font\relax}
\providecommand{\BIBforeignlanguage}[2]{{%
\expandafter\ifx\csname l@#1\endcsname\relax
\typeout{** WARNING: IEEEtran.bst: No hyphenation pattern has been}%
\typeout{** loaded for the language `#1'. Using the pattern for}%
\typeout{** the default language instead.}%
\else
\language=\csname l@#1\endcsname
\fi
#2}}
\providecommand{\BIBdecl}{\relax}
\BIBdecl

\bibitem{ref1}
Z.~Wu, S.~Pan, F.~Chen, G.~Long, C.~Zhang, and P.~S. Yu, ``A comprehensive survey on graph neural networks,'' \emph{IEEE Transactions on Neural Networks and Learning Systems}, vol.~32, no.~1, pp. 4--24, 2021.

\bibitem{DBLP:conf/iclr/MaharanaYB24}
A.~Maharana, P.~Yadav, and M.~Bansal, ``{D2} pruning: Message passing for balancing diversity {\&} difficulty in data pruning,'' in \emph{The Twelfth International Conference on Learning Representations, {ICLR} 2024, Vienna, Austria, May 7-11}, 2024.

\bibitem{zheng2022coverage}
H.~Zheng, R.~Liu, F.~Lai, and A.~Prakash, ``Coverage-centric coreset selection for high pruning rates,'' \emph{arXiv preprint arXiv:2210.15809}, 2022.

\bibitem{hong2021active}
S.~Hong and J.~Chae, ``Active learning with multiple kernels,'' \emph{IEEE transactions on neural networks and learning systems}, vol.~33, no.~7, pp. 2980--2994, 2021.

\bibitem{10537231}
A.~Moore, H.~Shim, J.~Zhu, and M.~Gong, ``Semi-supervised learning under general causal models,'' \emph{IEEE Transactions on Neural Networks and Learning Systems}, vol.~36, no.~4, pp. 7345--7356, 2025.

\bibitem{10138923}
Y.~Cui, W.~Deng, H.~Chen, and L.~Liu, ``Uncertainty-aware distillation for semi-supervised few-shot class-incremental learning,'' \emph{IEEE Transactions on Neural Networks and Learning Systems}, vol.~35, no.~10, pp. 14\,259--14\,272, 2024.

\bibitem{10239698}
W.~Li, Z.~Wang, X.~Yang, C.~Dong, P.~Tian, T.~Qin, J.~Huo, Y.~Shi, L.~Wang, Y.~Gao, and J.~Luo, ``Libfewshot: A comprehensive library for few-shot learning,'' \emph{IEEE Transactions on Pattern Analysis and Machine Intelligence}, vol.~45, no.~12, pp. 14\,938--14\,955, 2023.

\bibitem{author2025title}
\BIBentryALTinterwordspacing
R.~Tong, Y.~Liu, J.~Q. Shi, and D.~Gong, ``Coreset selection via reducible loss in continual learning,'' in \emph{Proceedings of the International Conference on Learning Representations (ICLR)}, 2025. [Online]. Available: \url{https://openreview.net/forum?id=mAztx8QO3B}
\BIBentrySTDinterwordspacing

\bibitem{zheng2025elfs}
H.~Zheng, E.~Tsai, Y.~Lu, J.~Sun, B.~R. Bartoldson, B.~Kailkhura, and A.~Prakash, ``Elfs: Label-free coreset selection with proxy training dynamics,'' in \emph{The Thirteenth International Conference on Learning Representations (ICLR)}, 2025.

\bibitem{gu2020efficient}
B.~Gu, Z.~Zhai, C.~Deng, and H.~Huang, ``Efficient active learning by querying discriminative and representative samples and fully exploiting unlabeled data,'' \emph{IEEE Transactions on Neural Networks and Learning Systems}, vol.~32, no.~9, pp. 4111--4122, 2020.

\bibitem{liu2021online}
S.~Liu, S.~Xue, J.~Wu, C.~Zhou, J.~Yang, Z.~Li, and J.~Cao, ``Online active learning for drifting data streams,'' \emph{IEEE Transactions on Neural Networks and Learning Systems}, vol.~34, no.~1, pp. 186--200, 2021.

\bibitem{7508942}
K.~Wang, D.~Zhang, Y.~Li, R.~Zhang, and L.~Lin, ``Cost-effective active learning for deep image classification,'' \emph{IEEE Transactions on Circuits and Systems for Video Technology}, vol.~27, no.~12, pp. 2591--2600, 2017.

\bibitem{9739135}
J.~W. Cho, D.-J. Kim, Y.~Jung, and I.~S. Kweon, ``Mcdal: Maximum classifier discrepancy for active learning,'' \emph{IEEE Transactions on Neural Networks and Learning Systems}, vol.~34, no.~11, pp. 8753--8763, 2023.

\bibitem{10251045}
X.~Nie, Z.~Deng, M.~He, M.~Fan, and Z.~Tang, ``Online active continual learning for robotic lifelong object recognition,'' \emph{IEEE Transactions on Neural Networks and Learning Systems}, vol.~35, no.~12, pp. 17\,790--17\,804, 2024.

\bibitem{zong2024bidirectional}
C.-C. Zong, Y.-W. Wang, K.-P. Ning, H.-B. Ye, and S.-J. Huang, ``Bidirectional uncertainty-based active learning for open-set annotation,'' in \emph{European Conference on Computer Vision}.\hskip 1em plus 0.5em minus 0.4em\relax Springer, 2024, pp. 127--143.

\bibitem{ning2022active}
K.-P. Ning, X.~Zhao, Y.~Li, and S.-J. Huang, ``Active learning for open-set annotation,'' in \emph{Proceedings of the IEEE/CVF Conference on Computer Vision and Pattern Recognition}, 2022, pp. 41--49.

\bibitem{li2024survey}
D.~Li, Z.~Wang, Y.~Chen, R.~Jiang, W.~Ding, and M.~Okumura, ``A survey on deep active learning: Recent advances and new frontiers,'' \emph{IEEE Transactions on Neural Networks and Learning Systems}, 2024.

\bibitem{pi2024unsupervised}
Y.~Pi, Y.~Shi, S.~Du, Y.~Huang, and S.~Wang, ``Unsupervised projected sample selector for active learning,'' \emph{IEEE Transactions on Big Data}, 2024.

\bibitem{ijcai2020p364}
\BIBentryALTinterwordspacing
C.~Li, H.~Ma, Z.~Kang, Y.~Yuan, X.-Y. Zhang, and G.~Wang, ``On deep unsupervised active learning,'' in \emph{Proceedings of the Twenty-Ninth International Joint Conference on Artificial Intelligence, {IJCAI-20}}, C.~Bessiere, Ed.\hskip 1em plus 0.5em minus 0.4em\relax International Joint Conferences on Artificial Intelligence Organization, 7 2020, pp. 2626--2632, main track. [Online]. Available: \url{https://doi.org/10.24963/ijcai.2020/364}
\BIBentrySTDinterwordspacing

\bibitem{zhang2011active}
L.~Zhang, C.~Chen, J.~Bu, D.~Cai, X.~He, and T.~S. Huang, ``Active learning based on locally linear reconstruction,'' \emph{IEEE Transactions on Pattern Analysis and Machine Intelligence}, vol.~33, no.~10, pp. 2026--2038, 2011.

\bibitem{li2018joint}
C.~Li, X.~Wang, W.~Dong, J.~Yan, Q.~Liu, and H.~Zha, ``Joint active learning with feature selection via cur matrix decomposition,'' \emph{IEEE transactions on pattern analysis and machine intelligence}, vol.~41, no.~6, pp. 1382--1396, 2018.

\bibitem{ma2022deep}
H.~Ma, C.~Li, X.~Shi, Y.~Yuan, and G.~Wang, ``Deep unsupervised active learning on learnable graphs,'' \emph{IEEE Transactions on Neural Networks and Learning Systems}, vol.~35, no.~2, pp. 2894--2900, 2022.

\bibitem{mindermann2022prioritized}
S.~Mindermann, J.~M. Brauner, M.~T. Razzak, M.~Sharma, A.~Kirsch, W.~Xu, B.~H{\"o}ltgen, A.~N. Gomez, A.~Morisot, S.~Farquhar \emph{et~al.}, ``Prioritized training on points that are learnable, worth learning, and not yet learnt,'' in \emph{International Conference on Machine Learning}.\hskip 1em plus 0.5em minus 0.4em\relax PMLR, 2022, pp. 15\,630--15\,649.

\bibitem{chen2022making}
L.~Chen, Y.~Bai, S.~Huang, Y.~Lu, B.~Wen, A.~L. Yuille, and Z.~Zhou, ``Making your first choice: To address cold start problem in vision active learning,'' \emph{arXiv preprint arXiv:2210.02442}, 2022.

\bibitem{10.5555/2540128.2540354}
F.~Nie, H.~Wang, H.~Huang, and C.~Ding, ``Early active learning via robust representation and structured sparsity,'' in \emph{Proceedings of the Twenty-Third International Joint Conference on Artificial Intelligence}, ser. IJCAI '13.\hskip 1em plus 0.5em minus 0.4em\relax AAAI Press, 2013, p. 1572–1578.

\bibitem{hu2013active}
Y.~Hu, D.~Zhang, Z.~Jin, D.~Cai, and X.~He, ``Active learning via neighborhood reconstruction,'' in \emph{Proc. IJCAI}, vol. 2013, 2013, pp. 1415--1421.

\bibitem{li2017active}
Q.~Li, X.~Shi, L.~Zhou, Z.~Bao, and Z.~Guo, ``Active learning via local structure reconstruction,'' \emph{Pattern Recognition Letters}, vol.~92, pp. 81--88, 2017.

\bibitem{li2021deep}
C.~Li, R.~Li, Y.~Yuan, G.~Wang, and D.~Xu, ``Deep unsupervised active learning via matrix sketching,'' \emph{IEEE Transactions on Image Processing}, vol.~30, pp. 9280--9293, 2021.

\bibitem{li2022structure}
C.~Li, H.~Ma, Y.~Yuan, G.~Wang, and D.~Xu, ``Structure guided deep neural network for unsupervised active learning,'' \emph{IEEE Transactions on Image Processing}, vol.~31, pp. 2767--2781, 2022.

\bibitem{sujit2023prioritizing}
S.~Sujit, S.~Nath, P.~Braga, and S.~Ebrahimi~Kahou, ``Prioritizing samples in reinforcement learning with reducible loss,'' \emph{Advances in Neural Information Processing Systems}, vol.~36, pp. 23\,237--23\,258, 2023.

\bibitem{evans2024bad}
T.~Evans, S.~Pathak, H.~Merzic, J.~Schwarz, R.~Tanno, and O.~J. Henaff, ``Bad students make great teachers: Active learning accelerates large-scale visual understanding,'' in \emph{European Conference on Computer Vision}.\hskip 1em plus 0.5em minus 0.4em\relax Springer, 2024, pp. 264--280.

\bibitem{tong2025coreset}
R.~Tong, Y.~Liu, J.~Q. Shi, and D.~Gong, ``Coreset selection via reducible loss in continual learning,'' in \emph{The Thirteenth International Conference on Learning Representations (ICLR)}, 2025.

\bibitem{margatina2021active}
K.~Margatina, G.~Vernikos, L.~Barrault, and N.~Aletras, ``Active learning by acquiring contrastive examples,'' \emph{arXiv preprint arXiv:2109.03764}, 2021.

\bibitem{maharana2023d2}
A.~Maharana, P.~Yadav, and M.~Bansal, ``D2 pruning: Message passing for balancing diversity and difficulty in data pruning,'' \emph{arXiv preprint arXiv:2310.07931}, 2023.

\bibitem{schuhmann2022laion}
C.~Schuhmann, R.~Beaumont, R.~Vencu, C.~Gordon, R.~Wightman, M.~Cherti, T.~Coombes, A.~Katta, C.~Mullis, M.~Wortsman \emph{et~al.}, ``Laion-5b: An open large-scale dataset for training next generation image-text models,'' \emph{Advances in neural information processing systems}, vol.~35, pp. 25\,278--25\,294, 2022.

\bibitem{hessel2021clipscore}
J.~Hessel, A.~Holtzman, M.~Forbes, R.~L. Bras, and Y.~Choi, ``Clipscore: A reference-free evaluation metric for image captioning,'' \emph{arXiv preprint arXiv:2104.08718}, 2021.

\bibitem{sorscher2022beyond}
B.~Sorscher, R.~Geirhos, S.~Shekhar, S.~Ganguli, and A.~Morcos, ``Beyond neural scaling laws: beating power law scaling via data pruning,'' \emph{Advances in Neural Information Processing Systems}, vol.~35, pp. 19\,523--19\,536, 2022.

\bibitem{mcqueen1967some}
J.~B. McQueen, ``Some methods of classification and analysis of multivariate observations,'' in \emph{Proc. of 5th Berkeley Symposium on Math. Stat. and Prob.}, 1967, pp. 281--297.

\bibitem{YangPLM}
Y.~{Yang}, Y.~{Wang}, Q.~M. {Jonathan Wu}, X.~{Lin}, and M.~{Liu}, ``Progressive learning machine: A new approach for general hybrid system approximation,'' \emph{IEEE Transactions on Neural Networks and Learning Systems}, vol.~26, no.~9, pp. 1855--1874, 2015.

\bibitem{yang2015extreme}
Y.~Yang and Q.~J. Wu, ``Extreme learning machine with subnetwork hidden nodes for regression and classification,'' \emph{IEEE transactions on cybernetics}, vol.~46, no.~12, pp. 2885--2898, 2015.

\bibitem{krizhevsky2009learning}
A.~Krizhevsky, G.~Hinton \emph{et~al.}, ``Learning multiple layers of features from tiny images,'' 2009.

\bibitem{deng2009imagenet}
J.~Deng, W.~Dong, R.~Socher, L.-J. Li, K.~Li, and L.~Fei-Fei, ``Imagenet: A large-scale hierarchical image database,'' in \emph{2009 IEEE conference on computer vision and pattern recognition}.\hskip 1em plus 0.5em minus 0.4em\relax Ieee, 2009, pp. 248--255.

\bibitem{papailiopoulos2014provable}
D.~Papailiopoulos, A.~Kyrillidis, and C.~Boutsidis, ``Provable deterministic leverage score sampling,'' in \emph{Proceedings of the 20th ACM SIGKDD international conference on Knowledge discovery and data mining}, 2014, pp. 997--1006.

\bibitem{cai2011manifold}
D.~Cai and X.~He, ``Manifold adaptive experimental design for text categorization,'' \emph{IEEE Transactions on Knowledge and Data Engineering}, vol.~24, no.~4, pp. 707--719, 2011.

\bibitem{li2024deep}
X.~Li, P.~Yang, Y.~Gu, X.~Zhan, T.~Wang, M.~Xu, and C.~Xu, ``Deep active learning with noise stability,'' in \emph{Proceedings of the AAAI Conference on Artificial Intelligence}, vol.~38, no.~12, 2024, pp. 13\,655--13\,663.

\bibitem{li2023bal}
J.~Li, P.~Chen, S.~Yu, S.~Liu, and J.~Jia, ``Bal: Balancing diversity and novelty for active learning,'' \emph{IEEE Transactions on Pattern Analysis and Machine Intelligence}, vol.~46, no.~5, pp. 3653--3664, 2023.

\bibitem{he2016deep}
K.~He, X.~Zhang, S.~Ren, and J.~Sun, ``Deep residual learning for image recognition,'' in \emph{Proceedings of the IEEE conference on computer vision and pattern recognition}, 2016, pp. 770--778.

\bibitem{yi2022pt4al}
J.~S.~K. Yi, M.~Seo, J.~Park, and D.-G. Choi, ``Pt4al: Using self-supervised pretext tasks for active learning,'' in \emph{European conference on computer vision}.\hskip 1em plus 0.5em minus 0.4em\relax Springer, 2022, pp. 596--612.

\end{thebibliography}

\begin{IEEEbiography}
[{\includegraphics[width=0.95in,height=1.25in]{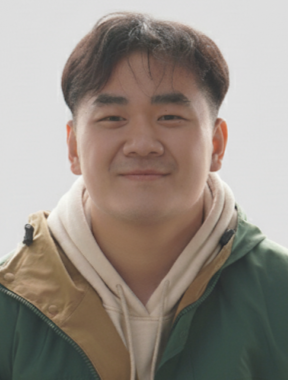}}]{Yuxi Liu}
 received the M.Eng degree in Control Science \& Engineering from the College of Electrical Engineering, Zhengzhou University, Zhengzhou, China, in 2022. 

He is currently a Ph.D. candidate in Electrical and Computer Engineering, Western University, Canada. His current research interests include artificial  neural networks, active learning and large language model. He is a reviewer for the IEEE Transactions
on Circuits and Systems for Video Technology.    
\end{IEEEbiography}

\begin{IEEEbiography}
[{\includegraphics[width=0.95in,height=1.25in]{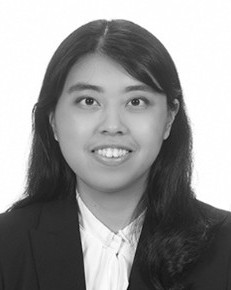}}]{Catherine Lalman}
 received B.Sc in Premedicine–Medicine with a minor in Mathematics from Pennsylvania State University and is an MD candidate at Sidney Kimmel Medical College, Thomas Jefferson University, Philadelphia, PA, USA. 
 
 Her research focuses on artificial intelligence, biomedical imaging, and transcriptomic data integration for ophthalmology and ocular disease. She has coauthored six manuscripts in journals and her work spans deep learning for optical coherence tomography image denoising, AI validation for glaucoma screening from fundus photography, supervised machine learning for gene signature identification, and integrative multi-omics analysis of fibrotic and regenerative pathways in the eye. She has been recognized with the Sigma Xi Grant-in-Aid of Research and the ARVO NES Award for research excellence.    
\end{IEEEbiography}

\begin{IEEEbiography}
[{\includegraphics[width=0.95in,height=1.25in]{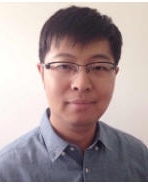}}]{Yimin Yang}
(S’10-M’13-SM’19) received the Ph.D. degree in Pattern Recognition and Intelligent System from the College of Electrical and Information Engineering, Hunan University, Changsha, China, in 2013. 

He is currently an Assistant Professor with the Department of Electrical and Computer Engineering, Western University, Canada. He is also a Faculty Affiliate Member in Vector Institute, Canada. From 2014 to 2018, he was a Post-Doctoral Fellow with the Department of Electrical and Computer Engineering at the University of Windsor, Canada. His current research interests are artificial neural networks, image processing, and robotics. 

Dr. Yang is an Associate Editor for the IEEE Transactions on Circuits and Systems for Video Technology, and the Journal of Neurocomputing.  He has been serving as a Reivewer for many international journals of his research field and a Program Committee Member of some international Conferences.  
\end{IEEEbiography}

\vfill

\end{document}